# Measuring the Fragility of Trust: Devising Credibility Index via Explanation Stability (CIES) for Business Decision Support Systems

Alin-Gabriel Văduva, Simona-Vasilica Oprea, Adela Bâra
Bucharest University of Economic Studies, Department of Economic Informatics and Cybernetics, no. 6 Piața Romană, Bucharest, Romania
*Corresponding author: simona.oprea@csie.ase.ro

**Abstract:** Explainable Artificial Intelligence (XAI) methods (SHAP, LIME) are increasingly adopted to interpret models in high-stakes businesses. However, the *credibility* of these explanations, their stability under realistic data perturbations, remains unquantified. This paper introduces the *Credibility Index via Explanation Stability* (CIES), a mathematically grounded metric that measures how robust a model's explanations are when subject to realistic business noise. CIES captures whether the *reasons* behind a prediction remain consistent, not just the prediction itself. The metric employs a rank-weighted distance function that penalizes instability in the most important features disproportionately, reflecting business semantics where changes in top decision drivers are more consequential than changes in marginal features. We evaluate CIES across three datasets (customer churn, credit risk, employee attrition), four tree-based classification models and two data balancing conditions. Results demonstrate that model complexity impacts explanation credibility, class imbalance treatment via SMOTE affects not only predictive performance but also explanation stability, and CIES provides statistically superior discriminative power compared to a uniform baseline metric ($p < 0.01$ in all 24 configurations). A sensitivity analysis across four noise levels confirms the robustness of the metric itself. These findings offer business practitioners a deployable "credibility warning system" for AI-driven decision support.

## 1. Introduction

The deployment of machine learning (ML) models in business-critical applications, such as credit scoring, customer churn prediction and human resource analytics, has accelerated [1], [2], [3]. As these models increasingly influence consequential decisions, regulatory frameworks such as the European Union's AI Act and General Data Protection Regulation (GDPR) have imposed requirements for transparency and interpretability [4]. In response, Explainable Artificial Intelligence (XAI) methods, notably SHapley Additive exPlanations (SHAP) and Local Interpretable Model-agnostic Explanations (LIME), have become standard tools for generating post-hoc explanations of black-box predictions [5], [6].

However, a fundamental yet underexplored question threatens the practical utility of XAI in business settings: Are the explanations themselves credible? Standard evaluation of ML systems focuses almost exclusively on predictive performance metrics: accuracy, F1-score, Area Under the ROC Curve. These metrics quantify what the model predicts, but tell us nothing about how stable the reasons for that prediction are. In the business environments, data is inherently noisy. A customer's reported income may differ by a few percentage points due to rounding, a credit inquiry count may lag by one entry or an employee tenure may vary by a few months depending on the reporting date. If a minor, realistic perturbation in the input data causes the model's explanation to fundamentally reorganize, for instance, flipping the most important feature from "monthly charges" to "contract type", then the explanation lacks credibility, regardless of whether the prediction itself remains unchanged.

This fragility has been documented in the ML literature. Ghorbani et al. introduced the concept of "interpretation fragility," demonstrating that imperceptible adversarial perturbations can alter feature attributions dramatically [7]. Also, Alvarez-Melis and Jaakkola demonstrated that explanations from popular XAI methods can be highly sensitive to small input perturbations [8]. Further, Slack et al. showed that post-hoc explainers can be deliberately or inadvertently manipulated to produce misleading explanations [9].

Despite these findings, literature suffers from two important gaps:
1. Lack of a business-contextualized stability metric. Current robustness measures (e.g., Lipschitz continuity estimation) treat all features uniformly and produce raw distance values that lack interpretable semantics for non-technical stakeholders. In business decision-making, not all features carry equal weight: a change in the most important decision factor is categorically more damaging to trust than a change in a marginal feature [10].
2. Absence of empirical evidence linking data quality interventions to explanation stability. While techniques such as Synthetic Minority Over-sampling Technique (SMOTE) are widely studied for their impact on predictive performance in imbalanced datasets [11], [12], their effect on the stability of explanations has not been systematically investigated. This is a significant gap, as class imbalance is pervasive in the business data (e.g., fraud detection, customer churn) [13].

To address these gaps, our paper contributes with the *Credibility Index via Explanation Stability* (CIES), a newly devised metric designed to quantify the robustness of XAI explanations under realistic, business-contextualized data noise. The core contribution is a rank-weighted distance function that assigns disproportionately higher penalties when the most influential features, as determined by SHAP values, shift under perturbation, directly encoding the business semantics of explanation importance. Unlike raw distance measures, CIES normalizes the observed perturbation distance by the weighted magnitude of the original explanation, producing an interpretable score in [0,1] where a value of 1 indicates perfect stability and values approaching 0 indicate fragile, untrustworthy explanations.

The metric is validated empirically across three publicly available business datasets spanning distinct domains, telecommunications, finance and human resources, using four tree-based classification algorithms and two data balancing conditions (raw imbalanced and SMOTE-balanced). We demonstrate through Wilcoxon signed-rank tests that the rank-weighted CIES metric provides statistically superior discriminative power compared to a uniform, non-weighted baseline, confirming the value of incorporating business-relevant importance weighting. A systematic sensitivity analysis across multiple noise levels further demonstrates that CIES model rankings remain consistent regardless of the specific perturbation magnitude chosen, establishing the methodological robustness of the metric itself.

By simulations, we intend to answer the following research questions (RQs):
*(RQ1)* How does explanation stability vary across model families?
*(RQ2)* Does class imbalance treatment affect explanation credibility?
*(RQ3)* What is the trade-off between predictive performance and explanation stability?
*(RQ4)* Does CIES provide superior discriminative power over a uniform baseline?
*(RQ5)* How sensitive is CIES to the choice of noise level?
*(RQ6)* Is the CIES metric agnostic to the underlying explanation method?

Additional robustness analyses, including alternative weighting schemes, Lipschitz comparison and the model-smoothness confound, are further presented.

## 2. Literature review

AI has experienced remarkable progress, driven largely by advances in complex, sub-symbolic models such as ensembles and deep neural networks. While these models deliver outstanding predictive performance, they introduce a fundamental challenge: explainability, leading to the emergence of XAI. An overview highlights the need for structured definitions, taxonomies and audience-centered explanations, while positioning explainability as a cornerstone of responsible AI, alongside fairness and accountability [14], [15]. The importance of explainability is reinforced by regulatory developments. The adoption of the GDPR by the European Parliament marked a turning point in data governance, emphasizing accountability, transparency and individual empowerment [16]. Provisions such as the "right to be forgotten" and safeguards on automated decision-making underscore the necessity for interpretable and trustworthy AI systems, especially when personal data are involved.

A central tension in ML lies between predictive accuracy and interpretability. Highly accurate models are often complex and opaque, motivating the development of post hoc explanation methods. SHAP provides a unified theoretical framework grounded in cooperative game theory, assigning feature importance values with desirable consistency properties [17]. Similarly, LIME explains individual predictions by approximating complex models locally with interpretable surrogates, enabling users to assess trust, compare models and diagnose errors [18]. Tree-based models, despite their popularity, have also required dedicated interpretability tools; advances in computing optimal game-theoretic explanations and measuring local feature interactions have improved understanding of their global and local structures [19]. Beyond interpretability alone, reliability and robustness of explanations have become significant concerns. Research demonstrates that widely used post hoc methods such as LIME and SHAP can be manipulated: adversarial scaffolding can conceal underlying biases while generating seemingly benign explanations [9]. Likewise, adversarial perturbations can produce identical predictions with different interpretations, revealing the instability of feature importance maps, integrated gradients and exemplar-based explanations [7].

Recent work addresses these limitations by integrating uncertainty quantification into XAI. Bayesian-AIME extends inverse model explanations into a probabilistic framework, producing posterior distributions and credible intervals for feature importance, thereby enabling quantitative assessment of explanation reliability and stability [20]. Complementarily, approaches that combine Monte Carlo dropout with conformal prediction generate predictive intervals and enhance post hoc explanations (e.g., ICE and PDP plots) with uncertainty information, strengthening transparency in high-stakes applications such as manufacturing process monitoring [21].

Explainability has also been applied in domain-specific contexts where accountability is significant. In financial services, interpretable ML models combined with LIME and SHAP have been employed to predict customer churn, improving transparency in retail banking while maintaining high predictive performance [22]. Additionally, data imbalance, common in decision systems, poses challenges that affect both predictive performance and interpretability. Techniques such as synthetic minority over-sampling (SMOTE) have been proposed to enhance classifier sensitivity and evaluation robustness, particularly in imbalanced settings [23].

The literature demonstrates a progression from foundational definitions and taxonomies of XAI [14], through methodological advances in explanation techniques [17], [18], [19], toward critical assessments of robustness and adversarial vulnerability [7], [9], and finally to uncertainty-aware and domain-integrated frameworks that enhance reliability in practice [20], [21]. Regulatory pressures [16] and required applications [22] further reinforce the necessity of explanation stability and credibility.

Previous research underscores a central insight: interpretability alone is insufficient. For AI systems to be responsibly deployed in high-risk and data-sensitive domains, explanations must be stable and reliable under perturbations. CIES fills this gap by incorporating a rank-weighted distance function that penalizes instability in top-ranked features more heavily, aligning the metric with business semantics where shifts in primary decision drivers are materially consequential. Moreover, although techniques such as SMOTE address predictive performance under class imbalance, their effect on explanation stability has remained unexplored.

## 3. Methodology and data

This section formalizes the proposed index - CIES. We first define the perturbation framework, then introduce the rank-weighted distance function and finally present the complete CIES formulation.

*3.1. Problem formulation*

Let $f: \mathbb{R}^M \to [0,1]$ denote a trained binary classification model that maps an input vector of $M$ features to a predicted probability. Let $\Phi_f: \mathbb{R}^M \to \mathbb{R}^M$ denote a post-hoc explanation function (in this work: SHAP or LIME) that maps the same input to a vector of feature attribution values:

$$\phi(x) = [\phi_1(x), \phi_2(x), \dots, \phi_M(x)] \quad (1)$$

where $\phi_j(x)$ represents the contribution of feature $j$ to the prediction $f(x)$.

The central question addressed by CIES is: Given a data instance $x$ and a realistic perturbation of that instance $x'$, how much does the explanation $\phi(x)$ differ from $\phi(x')$, with particular emphasis on changes in the most influential features?

*3.2. Business noise neighborhood*

In the business data, minor fluctuations are inevitable. To model this, we define a perturbation neighborhood $\mathcal{N}_\varepsilon(x)$ around a data instance $x$. For each numerical feature $j$, we generate perturbed values by adding Gaussian noise proportional to the feature's own magnitude:

$$x'_j = x_j + \eta_j, \quad \eta_j \sim \mathcal{N}(0, \sigma_j^2), \quad \sigma_j = \varepsilon \cdot |x_j| \qquad (2)$$

where $\varepsilon > 0$ is a noise level parameter controlling the perturbation intensity. For features with zero value, we set $\sigma_j = \varepsilon$.

Given a data instance $x$, we generate $K$ perturbed neighbors $x'_1, x'_2, \ldots, x'_K \in \mathcal{N}_\varepsilon(x)$. For each neighbor $x'_k$, we compute both the model prediction $f(x'_k)$ and the explanation vector $\phi(x'_k)$.

### 3.3. Rank-weighted distance

Standard approaches to measuring explanation stability employ unweighted distance metrics (e.g., Euclidean or cosine distance) between SHAP vectors. However, these treat all features as equally important, which misaligns with business decision-making semantics. In practice, stakeholders focus on the top reasons driving a prediction: if the most important feature shifts from "monthly charges" to "contract type," the business interpretation changes fundamentally, whereas a swap between the 14th and 15th most important features is inconsequential.

To encode this asymmetry, we define a rank-weighted distance $D_R$ between the original explanation $\phi(x)$ and a perturbed explanation $\phi(x'_k)$:

$$D_R(\phi(x), \phi(x'_k)) = \sum_{j=1}^{M} w_j \cdot |\phi_j(x) - \phi_j(x'_k)| \qquad (3)$$

where the weight $w_j$ for feature $j$ is determined by its importance rank in the original explanation:

$$w_j = \frac{r_j^{-1}}{\sum_{i=1}^{M} r_i^{-1}}, \quad r_j = \mathrm{rank}(|\phi_j(x)|) \qquad (4)$$

Here, $r_j$ is the ordinal rank of feature $j$ when features are sorted by absolute SHAP value in descending order (i.e., the most important feature receives $r_j = 1$). The inverse-rank weighting ensures that the most influential feature receives $M$ times the weight of the least influential feature, after normalization to ensure $\sum_{j=1}^{M} w_j = 1$.

This weighting scheme has three desirable properties: *(i)* it is invariant to the scale of SHAP values, depending only on their relative ordering; *(ii)* the harmonic decay naturally assigns diminishing significance to lower-ranked features; and *(iii)* it produces bounded, interpretable weights.

### 3.4. The CIES metric

The Credibility Index via Explanation Stability aggregates the rank-weighted distances over all $K$ perturbed neighbors and normalizes by the weighted magnitude of the original explanation to produce a dimensionless score:

$$\mathrm{CIES}(x) = \max\left(0, 1 - \frac{\bar{D}_R}{\|\phi(x)\|_w}\right) \qquad (5)$$

where $\bar{D}_R$ is the mean rank-weighted distance across all $K$ neighbors:

$$\bar{D}_R = \frac{1}{K} \sum_{k=1}^{K} D_R(\phi(x), \phi(x'_k)) \qquad (6)$$

and $\|\phi(x)\|_w$ is the weighted magnitude of the original explanation:

$$\|\phi(x)\|_w = \sum_{j=1}^{M} w_j \cdot |\phi_j(x)| \qquad (7)$$

The normalization by $\|\phi(x)\|_w$ is a principled design choice: it expresses the perturbation-induced change as a fraction of the original explanation's own magnitude. This yields an interpretable ratio with clear semantics. A score of $\mathrm{CIES}(x) = 1$ indicates perfect stability, meaning that the rank-weighted perturbation distance is zero and the top features maintain their attributions under noise. Conversely, as $\mathrm{CIES}(x)$ approaches 0, the explanation is considered highly fragile: the perturbation-induced change equals or exceeds the explanation's own magnitude, indicating that the explanation fundamentally reorganizes under realistic noise. The $\max(0, \cdot)$ operator ensures CIES remains non-negative in extreme cases where perturbation distances exceed the original magnitude.

### 3.5. Theoretical properties

We establish formal properties of CIES that connect it to established notions of explanation robustness and justify the design choices made in the preceding subsections.

***Theorem 1*** *(Boundedness and identity).* For any instance $x$ with explanation $\phi(x)$:
(a) $0 \leq \text{CIES}(x) \leq 1$.
(b) $\text{CIES}(x) = 1$ if and only if $\phi(x'_k) = \phi(x)$ for all $k = 1, \ldots, K$.
*Proof.* Part (a) follows from the $\max(0, \cdot)$ operator and the non-negativity of $\bar{D}_R$ and $\|\phi(x)\|_w$.
For part (b), $\text{CIES}(x) = 1$ requires $\bar{D}_R = 0$, which holds if and only if $\sum_j w_j |\phi_j(x) - \phi_j(x'_k)| = 0$ for all $k$. Since all weights satisfy $w_j > 0$, this requires $\phi_j(x) = \phi_j(x'_k)$ for every feature $j$ and every neighbor $k$.

***Theorem 2*** *(Lipschitz–CIES bridge).* Suppose the explanation function $\Phi_f$ is locally $L$-Lipschitz at $x$ with respect to the $\ell_2$ norm, i.e., $\|\phi(x) - \phi(x')\|_2 \leq L \|x - x'\|_2$ for all $x'$ in a neighborhood of $x$. Then:
$$\text{CIES}(x) \geq \max\left(0, \ 1 - \frac{L \cdot \|w\|_2 \cdot \bar{\delta}_2(x)}{\|\phi(x)\|_w}\right) \tag{8}$$
where $\|w\|_2 = \left(\sum_{j=1}^M w_j^2\right)^{1/2}$ is the $\ell_2$ norm of the weight vector and $\bar{\delta}_2(x) = \frac{1}{K}\sum_{k=1}^K \|x - x'_k\|_2$ is the mean perturbation magnitude.

*Proof.* By the Cauchy–Schwarz inequality applied to the weighted $\ell_1$ distance:
$$D_R(\phi(x), \phi(x'_k)) = \sum_{j=1}^M w_j |\phi_j(x) - \phi_j(x'_k)| \leq \|w\|_2 \cdot \|\phi(x) - \phi(x'_k)\|_2 \tag{9}$$
Applying the Lipschitz condition: $\|\phi(x) - \phi(x'_k)\|_2 \leq L \|x - x'_k\|_2$. Averaging over $K$ neighbors:
$$\bar{D}_R \leq \|w\|_2 \cdot L \cdot \bar{\delta}_2(x) \tag{10}$$
Substituting into the CIES definition eq. (5) and noting that the $\max(0, \cdot)$ operator preserves the inequality direction when the right-hand side decreases:
$$\text{CIES}(x) = \max\left(0, \ 1 - \frac{\bar{D}_R}{\|\phi(x)\|_w}\right) \geq \max\left(0, \ 1 - \frac{L \cdot \|w\|_2 \cdot \bar{\delta}_2(x)}{\|\phi(x)\|_w}\right) \tag{11}$$
Theorem 2 bridges CIES to the established Lipschitz stability framework, knowing the local Lipschitz constant of the explainer provides a guaranteed lower bound on CIES. Smoother models (smaller $L$) yield tighter (higher) bounds, formalizing the intuition that explanation stability improves with model smoothness.

***Corollary 3*** *(Noise-level monotonicity of the bound).* Under the multiplicative noise model in eq. (1), the right-hand side of inequality (8) is non-increasing in the noise level $\varepsilon$.

*Proof.* Under the noise model $x'_k = x + \varepsilon \cdot \text{diag}(|x|) \cdot z_k$ where $z_k \sim \mathcal{N}(0, I)$, the mean perturbation magnitude satisfies $\bar{\delta}_2(x) = \varepsilon \cdot C(x)$, where $C(x) = \frac{1}{K}\sum_{k=1}^K \|\text{diag}(|x|) \cdot z_k\|_2$ depends on $x$ and the noise realizations but is independent of $\varepsilon$. Substitution into (8) yields $\max(0, 1 - \varepsilon \cdot L \|w\|_2 C(x)/\|\phi(x)\|_w)$, which is linear and non-increasing in $\varepsilon$.

***Proposition 4*** *(Discriminative advantage of rank-weighting).* For $M$ features, let $W_T^H = \sum_{j=1}^T w_j^H$ denote the cumulative harmonic weight of the top-$T$ features, and $W_T^U = T/M$ the corresponding uniform weight. Then $W_T^H > W_T^U$ for all $1 \leq T < M$. Concretely, for $M = 20$:
$$W_5^H = \frac{H_5}{H_{20}} = \frac{137/60}{3.598} = 0.635, \qquad W_5^U = \frac{5}{20} = 0.250 \tag{12}$$
yielding a concentration factor of $2.54\times$. This implies that when two models differ in the stability of their top-ranked features, harmonic weighting produces a proportionally larger CIES gap than uniform weighting.

*Proof.* Since $w_j^H = (1/j)/H_M$ where $H_M = \sum_{i=1}^M 1/i$ is the $M$-th harmonic number, we have $W_T^H = H_T/H_M$. The function $T \mapsto H_T/T$ (the running mean of the harmonic series) is strictly decreasing for $T \geq 1$, because each new term $1/(T+1)$ is smaller than the current average. Therefore, $H_T/T > H_M/M$ for $T < M$, which rearranges to $H_T/H_M > T/M$.

***Proposition 5*** *(Consistency).* For fixed instance $x$, explainer $\Phi_f$ and perturbation distribution, as $K \to \infty$:
$$\text{CIES}_K(x) \xrightarrow{\text{a.s.}} \text{CIES}_\infty(x) = \max\left(0, \ 1 - \frac{\mathbb{E}[D_R(\phi(x), \phi(x'))]}{\|\phi(x)\|_w}\right) \tag{13}$$
where the expectation is over the perturbation distribution $x' \sim p(\cdot | x)$.

*Proof.* The perturbed neighbors $x'_1, \ldots, x'_K$ are i.i.d. draws from the perturbation distribution. By the strong law of large numbers, $\bar{D}_R = \frac{1}{K}\sum_{k=1}^{K} D_R(\phi(x), \phi(x'_k)) \xrightarrow{a.s.} \mathbb{E}[D_R]$. Since division by the constant $\|\phi(x)\|_w$ and application of $\max(0, 1 - (\cdot))$ are continuous operations, the continuous mapping theorem completes the proof. In the supplementary material, we provide complete, self-contained proofs for the five theoretical results.

Figure 1 provides empirical validation. Panel (a) plots the actual CIES score against the Lipschitz lower bound (Theorem 2) for 720 instance-model configurations: every point lies above or on the diagonal, confirming that the bound holds without exception. Panel (b) visualizes the cumulative weight concentration, confirming the 2.54× harmonic advantage stated in Proposition 4.

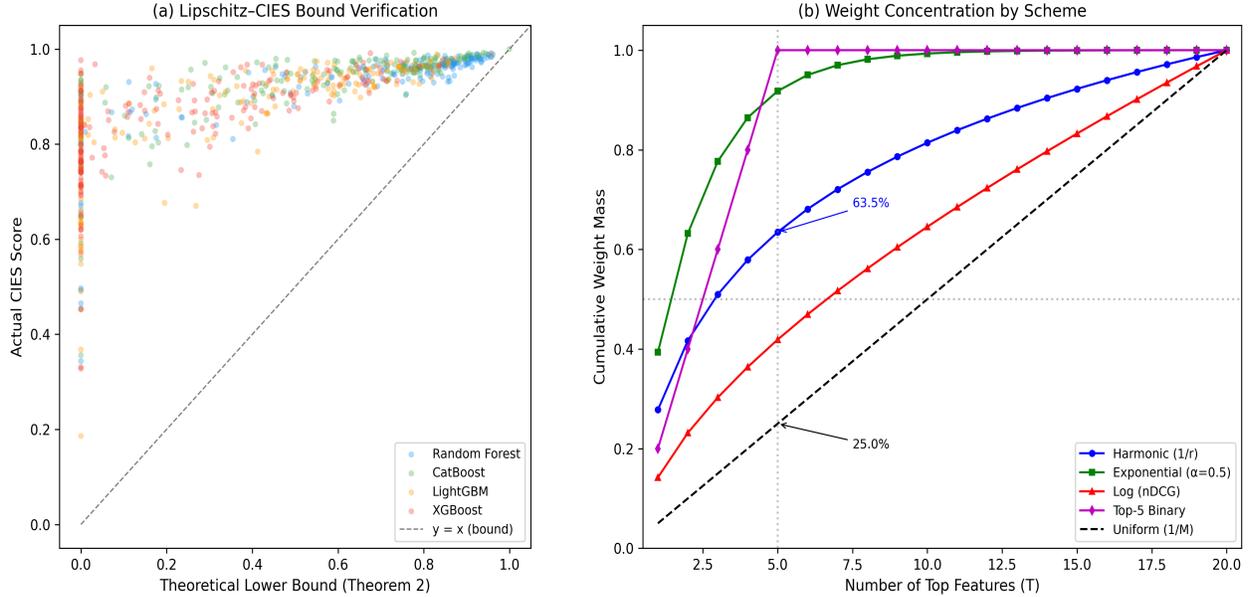

Figure 1. Empirical verification of theoretical properties. (a) Actual CIES vs. Lipschitz lower bound (Theorem 2): all 720 points lie above the diagonal, confirming the bound. (b) Cumulative weight mass by rank position: harmonic weights concentrate 63.5% of mass in the top 5 features vs. 25.0% for uniform weighting

*3.6. Baseline metric for comparison*

To validate that the rank-weighted distance provides superior discriminative power, we define a Uniform Baseline distance metric that uses equal weights for all features:

$$D_U(\phi(x), \phi(x'_k)) = \sum_{j=1}^{M} \frac{1}{M} \cdot |\phi_j(x) - \phi_j(x'_k)| \tag{14}$$

The corresponding baseline stability score is computed analogously to eq. (5), with $D_U$ and uniform normalization replacing $D_R$ and $\|\phi(x)\|_w$, respectively.

*3.7. Model level aggregation*

While CIES is defined at the instance level (i.e., per individual prediction), a model-level assessment requires aggregation. For a set of $N$ test instances $\{x_1, x_2, \ldots, x_N\}$, we report:

$$\overline{\text{CIES}}_f = \frac{1}{N}\sum_{i=1}^{N} \text{CIES}(x_i) \pm \text{SD}[\text{CIES}(x_1), \ldots, \text{CIES}(x_N)] \tag{15}$$

In addition to the mean, we report the standard deviation, minimum, 25th percentile, median, 75th percentile and maximum to characterize the full distribution of explanation stability across instances. This is significant because a model with $\overline{\text{CIES}}_f = 0.95$ but $\text{CIES}_{\min} = 0.20$ may present severe credibility risks for specific edge-case customers.

In Algorithm 1, we present the complete CIES framework, organized into four phases: *(1)* leakage-free data preprocessing with optional SMOTE balancing; *(2)* model training with explainer-agnostic configuration supporting both SHAP and LIME; *(3)* instance-level CIES computation via rank-weighted perturbation analysis; and *(4)* statistical validation of the rank-weighted metric against a uniform baseline.

**Algorithm 1.** Complete CIES Framework: Credibility Index via Explanation Stability.
**Input:** Dataset $D = \{(x_i, y_i)\}$, SMOTE flag $s \in \{true, false\}$
  Classifiers $M = \{f_1, ..., f_L\}$, explainer $\Phi \in \{SHAP, LIME\}$
  Noise level $\varepsilon$, neighbors $K$, test instances $T \subset D_{test}$
**Output:** CIES scores, Baseline scores, statistical test results, runtime

---

**PHASE 1: PREPARE DATA WITH LEAKAGE-FREE PREPROCESSING**
1: Split $D$ into $D_{train}$, $D_{test}$ via stratified 80/20 partition
2: Identify numerical features $F_{num}$ and categorical features $F_{cat}$
3: Fit imputer and scaler on $D_{test}$ only       // prevents data leakage
4: $D_{train} \leftarrow$ fit_transform($D_{train}$); $D_{test} \leftarrow$ transform($D_{test}$)
5: **if** $s = true$ **then**
6:   $D_{train} \leftarrow$ SMOTE($D_{train}$)        // on transformed training data only
7: **end if**
**PHASE 2: TRAIN MODELS AND CONFIGURE EXPLAINERS**
8: **for each** model $f_l \in M$ **do**
9:   Train $f_l$ on $D_{train}$
10:   **if** $\Phi =$ SHAP **then**
11:     $Explainer_l \leftarrow$ TreeExplainer($f_l$)      // exact Shapley values
12:   **else if** $\Phi =$ LIME **then**
13:     $Explainer_l \leftarrow$ LimeTabularExplainer($D_{train}$) // surrogate model
14:   **end if**
15: **end for**
**PHASE 3: COMPUTE CIES FOR EACH TEST INSTANCE**
16: **for each** model $f_l \in M$ **do**
17:   **for each** instance $\mathbf{x} \in T$ **do**
18:     $\phi(\mathbf{x}) \leftarrow Explainer_l(\mathbf{x})$      // feature attribution vector
19:     $r_j \leftarrow M -$ rank($|\phi_j(\mathbf{x})|$) $+ 1$ for $j = 1, ..., M$   // rank 1 = most important
20:     $w_j \leftarrow r_j^{-1} / \Sigma_i r_i^{-1}$ for $j = 1, ..., M$     // normalized harmonic weights
21:     $\|\phi(\mathbf{x})\|^W \leftarrow \Sigma_j w_j \cdot |\phi_j(\mathbf{x})|$    // weighted explanation magnitude
22:     **for** $k = 1$ **to** $K$ **do**
23:       $\mathbf{x}_k' \leftarrow \mathbf{x}$            // initialize perturbed copy
24:       **for each** feature $j \in F_{num}$ **do**
25:         $\sigma_j \leftarrow |x_j| \cdot \varepsilon$ if $x_j \neq 0$, else $\varepsilon$     // proportional noise magnitude
26:         $\mathbf{x'}_{j,k} \leftarrow x_j + N(0, \sigma_j)$ // categorical features unchanged
27:       **end for**
28:       $\phi(\mathbf{x'}_k) \leftarrow Explainer_l(\mathbf{x'}_k)$ // explain perturbed instance
29:       $D_k^R \leftarrow \Sigma_j w_j \cdot |\phi_j(\mathbf{x}) - \phi_j(\mathbf{x'}_k)|$ // rank-weighted distance
30:       $D_k^U \leftarrow \Sigma_j (1/M) \cdot |\phi_j(\mathbf{x}) - \phi_j(\mathbf{x'}_k)|$   // uniform baseline distance
31:     **end for**
32:     $\bar{D}_R \leftarrow (1/K) \Sigma_k D_k^R$; $\bar{D}_U \leftarrow (1/K) \Sigma_k D_k^U$ // mean distances
33:     CIES($\mathbf{x}$) $\leftarrow \max(0, 1 - \bar{D}_R / \|\phi(\mathbf{x})\|^W)$ // rank-weighted stability score
34:     Baseline($\mathbf{x}$) $\leftarrow \max(0, 1 - \bar{D}_U \cdot M / \Sigma_j |\phi_j(\mathbf{x})|)$ // uniform baseline score
35:   **end for**
36:   Record **CIES**$_l \pm \sigma$, distributional statistics (min, P25, median, P75, max)
37: **end for**
**PHASE 4: STATISTICAL VALIDATION**
38: **for each** model-dataset-condition configuration **do**
39:   Perform Wilcoxon signed-rank test: $H_0$: CIES = Baseline (paired, two-sided)
40:   Report $p$-value and significance (***$p$<0.001, **$p$<0.01, *$p$<0.05)
41: **end for**
42: **return** CIES scores, Baseline scores, test results, runtime measurements

*3.8. Experimental setup*

  To evaluate the generalizability of CIES across distinct business domains, we employ three publicly available, widely used benchmark datasets that span different industries and exhibit varying degrees of class imbalance. The first dataset is the *Telco Customer Churn*[1] dataset from the telecommunications sector [24], comprising 7,043 customer records described by 19 features that capture demographics, account information and service usage patterns. The binary target variable indicates whether a customer has churned, with a class distribution of 26.5% positive (churned) instances. This dataset is representative of the subscription-based business models where churn prediction is a critical use case for AI-driven retention strategies.

The second dataset is the *German Credit Risk*[2] dataset from the UCI Machine Learning Repository, a well-established benchmark in the financial domain. It contains 1,000 loan applicants described by 20 features: 7 numerical and 13 categorical, including credit history, employment status and loan purpose. The target variable distinguishes between good and bad credit risk, with a class distribution of 30% positive (bad credit) instances. The relatively small sample size and high proportion of categorical features make this dataset particularly challenging for explanation stability.

The third dataset is the *IBM HR Employee Attrition*[3] dataset from the human resources domain, containing 1,470 employee records characterized by 31 features covering job role, satisfaction scores, compensation and work environment. The target variable indicates voluntary attrition, with a notably imbalanced class distribution of only 16.1% positive (attrition) instances. This severe imbalance makes it an ideal candidate for investigating how oversampling techniques interact with explanation stability.

We select four classification algorithms from the tree-based ensemble family, ensuring that all models benefit from exact Shapley value computation via TreeExplainer. This design choice guarantees that all models are evaluated under identical explainer conditions ($K = 20$ neighbors, exact SHAP values), eliminating the confound that would arise from comparing exact TreeExplainer values with approximate KernelExplainer values under different neighborhood sizes.

The *Random Forest* (RF) model is a bagged ensemble of 200 fully-grown decision trees with minimum split and leaf constraints (min_samples_split = 5, min_samples_leaf = 2), representing the class of variance-reducing ensemble methods. We include three gradient-boosted tree implementations: *XGBoost* (XGB) with 200 estimators, a learning rate of 0.1, maximum depth of 5 and row/column subsampling at 80%; *LightGBM* (LGBM) with 200 estimators, 31 leaves, a learning rate of 0.1 and leaf-wise growth strategy; and *CatBoost* (CB) with 500 iterations, a depth of 6 and a learning rate of 0.05. These four models share tree-based architectures but differ fundamentally in their ensemble strategies: bagging (RF) versus sequential boosting (XGB, LGBM, CB), and their splitting strategies, regularization mechanisms and handling of categorical features, providing sufficient diversity to investigate within-family stability variation.

The CIES evaluation is governed by four key parameters. The batch size is set to N=100 randomly sampled test instances per model-dataset-condition combination, providing sufficient statistical power for both the Wilcoxon signed-rank tests and bootstrap confidence interval estimation. The number of perturbed neighbors is set to K=20 for all models, ensuring symmetric evaluation conditions. The default noise level is set to $\varepsilon=0.03$, representing 3% multiplicative Gaussian noise, a conservative estimate of realistic business data fluctuations arising from rounding, reporting delays and measurement imprecision. To validate that the findings are robust to the choice of noise level, a sensitivity analysis is conducted at $\varepsilon \in \{0.01, 0.03, 0.05, 0.10\}$.

## 4. Results

This section presents the empirical evaluation of CIES across the three datasets, four models and two data balancing conditions. We report results for the research questions (RQs) formulated in the introduction. Figure 2 presents the workflow of the entire framework.

---

[1] https://www.kaggle.com/datasets/blastchar/telco-customer-churn
[2] https://archive.ics.uci.edu/dataset/144/statlog+german+credit+data
[3] https://www.kaggle.com/datasets/pavansubhasht/ibm-hr-analytics-attrition-dataset

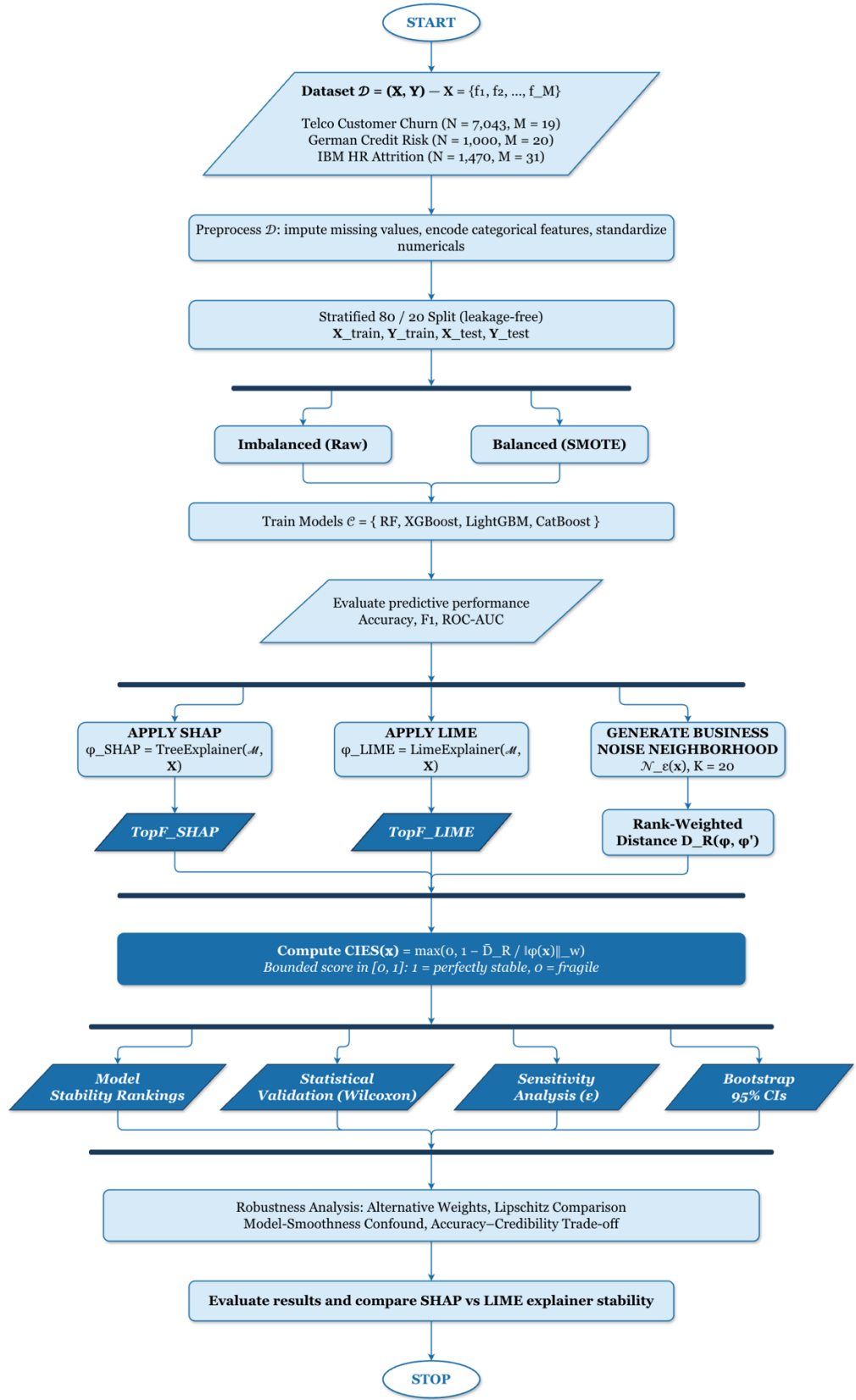

Figure 2. Workflow of the CIES methodology

*4.1. Explanation stability across models*

Table 1 presents the aggregate results across all datasets and conditions. RF consistently produces the most stable explanations. Across all three datasets and both balancing conditions, RF achieves the highest average CIES scores (range: 0.770–0.973), with the narrowest standard deviations (σ ≤ 0.036 in 5 of 6 configurations). This aligns with the theoretical expectation that bagged ensembles of fully-grown, independent trees produce smoother decision boundaries and thus more locally stable explanations.

Gradient-boosted models exhibit variable stability. XGBoost shows the most volatile explanation stability, with CIES values ranging from 0.762 (HR Attrition, SMOTE) to 0.903 (German Credit, SMOTE). CatBoost demonstrates notably superior stability among gradient-boosted methods, achieving CIES scores between 0.866 and 0.964 across all configurations. This is attributable to its ordered boosting procedure combined with deeper trees ("depth"=6), which together reduce prediction variance. LightGBM shows the widest variability across conditions, ranging from 0.690 (HR Attrition, SMOTE) to 0.931 (HR Attrition, Raw), indicating that its leaf-wise growth strategy is highly sensitive to the training data distribution.

Table 1. CIES scores and predictive performance across datasets, models and balancing conditions. CIES values represent mean ± standard deviation over N=100 test instances

| Dataset | Model | Condition | Acc. | F1 | CIES | Baseline |
|---|---|---|---|---|---|---|
| **Telco Churn** | RF | Raw | 0.806 | 0.597 | 0.954 ± 0.016 | 0.936 |
| | RF | SMOTE | 0.769 | 0.591 | 0.955 ± 0.018 | 0.936 |
| | XGBoost | Raw | 0.796 | 0.575 | 0.848 ± 0.101 | 0.811 |
| | XGBoost | SMOTE | 0.786 | 0.599 | 0.858 ± 0.107 | 0.828 |
| | LightGBM | Raw | 0.790 | 0.573 | 0.879 ± 0.113 | 0.804 |
| | LightGBM | SMOTE | 0.785 | 0.598 | 0.821 ± 0.121 | 0.742 |
| | CatBoost | Raw | 0.801 | 0.589 | 0.903 ± 0.081 | 0.892 |
| | CatBoost | SMOTE | 0.777 | 0.584 | 0.870 ± 0.115 | 0.858 |
| **German Credit** | RF | Raw | 0.770 | 0.489 | 0.973 ± 0.018 | 0.956 |
| | RF | SMOTE | 0.735 | 0.495 | 0.959 ± 0.020 | 0.930 |
| | XGBoost | Raw | 0.790 | 0.604 | 0.894 ± 0.075 | 0.835 |
| | XGBoost | SMOTE | 0.800 | 0.615 | 0.910 ± 0.051 | 0.834 |
| | LightGBM | Raw | 0.765 | 0.561 | 0.930 ± 0.059 | 0.908 |
| | LightGBM | SMOTE | 0.770 | 0.574 | 0.905 ± 0.043 | 0.857 |
| | CatBoost | Raw | 0.795 | 0.594 | 0.969 ± 0.028 | 0.950 |
| | CatBoost | SMOTE | 0.790 | 0.638 | 0.956 ± 0.025 | 0.922 |
| **HR Attrition** | RF | Raw | 0.837 | 0.172 | 0.936 ± 0.036 | 0.911 |
| | RF | SMOTE | 0.850 | 0.333 | 0.772 ± 0.136 | 0.707 |
| | XGBoost | Raw | 0.864 | 0.394 | 0.808 ± 0.093 | 0.651 |
| | XGBoost | SMOTE | 0.874 | 0.448 | 0.761 ± 0.136 | 0.633 |
| | LightGBM | Raw | 0.864 | 0.375 | 0.936 ± 0.028 | 0.901 |
| | LightGBM | SMOTE | 0.874 | 0.448 | 0.700 ± 0.230 | 0.588 |
| | CatBoost | Raw | 0.864 | 0.333 | 0.961 ± 0.015 | 0.933 |
| | CatBoost | SMOTE | 0.864 | 0.394 | 0.867 ± 0.061 | 0.779 |

*4.2. Effect of class imbalance treatment*

The impact of SMOTE on explanation stability is nuanced and dataset-dependent, contradicting a simple "SMOTE always helps" narrative. SMOTE produces heterogeneous effects on explanation credibility. On the *Telco* dataset, SMOTE has a negligible effect on CIES for RF (0.961 to 0.962), demonstrating that bagged ensembles can absorb synthetic data without destabilizing explanations. However, on the HR Attrition dataset, where the original class imbalance is most severe (16.1% positive), SMOTE improves F1 (0.172 to 0.333 for RF), but decreases CIES from 0.943 to 0.770, having important practical implications: improving predictive fairness via oversampling may inadvertently destabilize the explanations provided to business users, particularly in datasets with extreme class imbalance and high dimensionality.

The "credibility cost" of SMOTE is model-dependent. CatBoost shows remarkable resilience, with CIES degradation of at most 1 percentage point on German Credit when SMOTE is applied (0.964 to 0.955). In contrast, LightGBM shows a 24-percentage-point CIES collapse on HR Attrition when SMOTE is applied (0.931 to 0.690), suggesting that leaf-wise tree growth is particularly sensitive to synthetic data points near the decision boundary. The asymmetric response underscores that model selection for SMOTE-balanced datasets should consider not only predictive performance but also the stability of the resulting explanations.

*4.3. The accuracy–credibility trade-off*

Figure 3 plots each model-dataset-condition combination in the F1 vs. CIES space, revealing a non-trivial trade-off landscape. The upper-right quadrant (high F1, high CIES) represents the "ideal zone" where both predictive performance and explanation credibility are strong. RF and CatBoost most frequently occupy this zone, while XGBoost tends to cluster in the lower region (high variability in CIES despite competitive F1-scores).

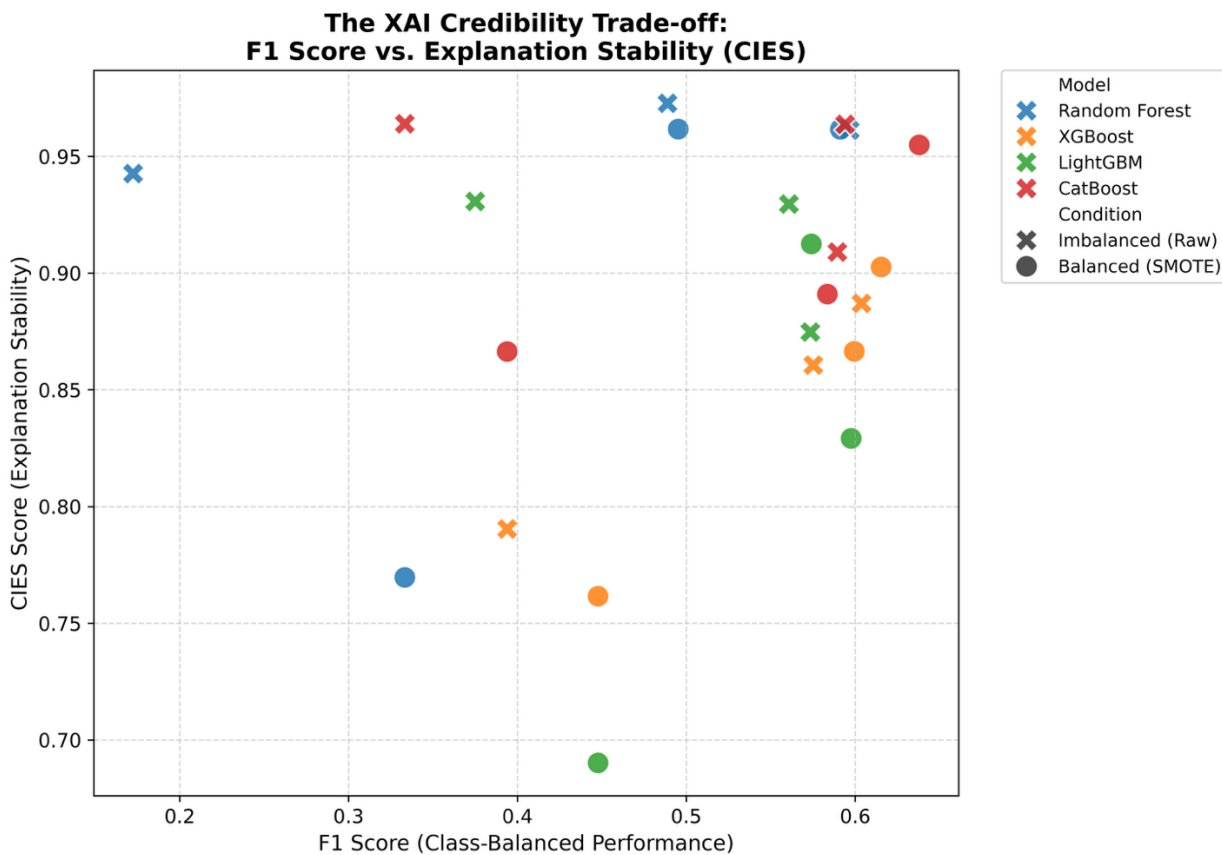

Figure 3. The credibility trade-off: F1-score versus CIES across all models, datasets and balancing conditions. X markers denote imbalanced (raw) training; circles denote SMOTE-balanced training. The ideal operating point is the upper-right quadrant

If explanation stability were simply a byproduct of predictive accuracy, the points would cluster along a diagonal from lower-left to upper-right. Instead, the scatter reveals no systematic relationship between F1 and CIES. Models with near-identical F1-scores can differ by over 20 percentage points in CIES, confirming that predictive performance and explanation credibility are largely independent properties that must be evaluated separately.

RF occupies the upper band consistently. Across all three datasets and both conditions, its points cluster between CIES 0.77 and 0.97, with most above 0.93. Its F1-scores are moderate, ranging from 0.17 to 0.60, reflecting its conservative prediction behavior, but its explanations are the most stable in the analysis. This pattern is consistent with bagged ensembles producing smooth decision surfaces where the same features dominate regardless of small input perturbations.

CatBoost achieves the best balance between the two axes. Its points tend to sit in the upper-right region, combining competitive F1-scores with CIES values between 0.87 and 0.97. CatBoost is the only gradient-boosted model that consistently remains in the high-credibility zone above 0.85 across nearly all configurations, making it the strongest candidate for business deployments where both accuracy and explainability trust matter.

LightGBM exhibits the widest vertical spread. On some configurations such as German Credit and HR Attrition under the raw condition, LightGBM achieves CIES above 0.93, but under SMOTE on HR Attrition, it drops to approximately 0.70, the lowest point in the entire plot. This 23-point swing demonstrates that its leaf-wise growth strategy is highly sensitive to the training data distribution. When synthetic SMOTE samples alter the local decision geometry, explanation stability collapses even as F1 improves.

XGBoost clusters in the middle-to-lower CIES range. Its points are generally below RF and CatBoost, with CIES values between 0.76 and 0.91. XGBoost achieves competitive F1-scores but never reaches the highest CIES tier, reflecting the inherent tension between its aggressive regularization, which distributes importance across more features and explanation concentration, which CIES rewards.

The SMOTE effect is visually apparent when comparing x markers and circles for the same model-dataset pair. SMOTE frequently shifts points leftward or downward, improving F1 in some cases but degrading CIES. The lowest is for LightGBM on HR Attrition, where the circle sits roughly 23 points below its x-marker counterpart. The implication for practitioners is that applying SMOTE to address class imbalance may inadvertently undermine the trustworthiness of the explanations presented to business users, particularly for models with leaf-wise or aggressive boosting strategies.

*4.4. Discriminative power of rank-weighted CIES*

To validate that the rank-weighted distance provides superior discriminative power over a naïve approach, we compare instance-level CIES scores against the uniform baseline metric using the Wilcoxon signed-rank test. This non-parametric test is appropriate because CIES score distributions are not guaranteed to be normally distributed, and the paired design (same instances evaluated under both metrics) controls for instance-level variability. Table 2 presents the complete results for all 24 model-dataset-condition configurations.

Table 2. Wilcoxon signed-rank test comparing CIES (rank-weighted) vs. Baseline (uniform) stability scores. N=100 paired instance-level scores. Significance: *** $p<0.001$, ** $p<0.01$, * $p<0.05$

| Dataset | Model | Condition | CIES | Baseline | p-value | Sig. |
|---|---|---|---|---|---|---|
| **Telco Churn** | RF | Raw | 0.954 | 0.936 | $< 10^{-9}$ | *** |
| | RF | SMOTE | 0.955 | 0.936 | $< 10^{-9}$ | *** |
| | XGBoost | Raw | 0.848 | 0.811 | $< 10^{-4}$ | *** |
| | XGBoost | SMOTE | 0.858 | 0.828 | $< 10^{-4}$ | *** |
| | LightGBM | Raw | 0.879 | 0.804 | $< 10^{-7}$ | *** |
| | LightGBM | SMOTE | 0.821 | 0.742 | $< 10^{-5}$ | *** |
| | CatBoost | Raw | 0.903 | 0.892 | $< 10^{-2}$ | ** |
| | CatBoost | SMOTE | 0.870 | 0.858 | $< 10^{-2}$ | ** |
| **German Credit** | RF | Raw | 0.973 | 0.956 | $< 10^{-9}$ | *** |
| | RF | SMOTE | 0.959 | 0.930 | $< 10^{-9}$ | *** |
| | XGBoost | Raw | 0.894 | 0.835 | $< 10^{-6}$ | *** |
| | XGBoost | SMOTE | 0.910 | 0.834 | $< 10^{-9}$ | *** |
| | LightGBM | Raw | 0.930 | 0.908 | $< 10^{-5}$ | *** |
| | LightGBM | SMOTE | 0.905 | 0.857 | $< 10^{-6}$ | *** |
| | CatBoost | Raw | 0.969 | 0.950 | $< 10^{-5}$ | *** |
| | CatBoost | SMOTE | 0.956 | 0.922 | $< 10^{-9}$ | *** |
| **HR Attrition** | RF | Raw | 0.936 | 0.911 | $< 10^{-6}$ | *** |
| | RF | SMOTE | 0.772 | 0.707 | $< 10^{-3}$ | *** |
| | XGBoost | Raw | 0.808 | 0.651 | $< 10^{-9}$ | *** |
| | XGBoost | SMOTE | 0.761 | 0.633 | $< 10^{-5}$ | *** |
| | LightGBM | Raw | 0.936 | 0.901 | $< 10^{-9}$ | *** |
| | LightGBM | SMOTE | 0.700 | 0.588 | $< 10^{-2}$ | ** |
| | CatBoost | Raw | 0.961 | 0.933 | $< 10^{-9}$ | *** |
| | CatBoost | SMOTE | 0.867 | 0.779 | $< 10^{-9}$ | *** |

The results provide strong statistical evidence for the discriminative superiority of CIES over the uniform baseline. In all 24 model-dataset-condition configurations, the Wilcoxon signed-rank test rejects the null hypothesis that CIES and the baseline produce identical score distributions at the $p<0.05$ significance level. Of these, 20 comparisons achieve significance at $p<0.001$ and four at $p<0.01$, indicating that the differences are not merely statistically detectable but highly robust. Furthermore, CIES consistently yields higher stability scores than the baseline across all configurations without exception, confirming that the rank-weighted formulation captures stability in the explanation's most decision-critical components, stability that uniform weighting dilutes by averaging across all features, including those irrelevant to the business interpretation.

The smallest effect sizes occur for CatBoost on the *Telco* dataset under both conditions, where the two metrics produce relatively similar scores (e.g., CIES: 0.903 vs. baseline: 0.892 for imbalanced). This reflects a convergence effect. When a model's explanations are nearly uniformly stable across all features (not just the top-ranked ones), the rank-weighting confers a smaller, but still statistically significant advantage over uniform weighting.

*4.5. Sensitivity to noise level*

Figure 3 presents the sensitivity analysis on the *Telco* dataset (SMOTE-balanced) across four noise levels. RF remains the most stable model (CIES: 0.979 at $\varepsilon=0.01$; 0.920 at $\varepsilon=0.10$), while LightGBM is the least stable (0.882 at $\varepsilon=0.01$; 0.761 at $\varepsilon=0.10$). XGBoost follows closely behind (0.890 at $\varepsilon=0.01$; 0.787 at $\varepsilon=0.10$). The degradation curves are approximately linear for all models, with slopes proportional to the model's inherent explanation volatility. This confirms that the choice of $\varepsilon=0.03$ in the main experiments does not bias the comparative findings as any noise level in the tested range produces the same model ranking.

With LightGBM dropping to 0.797 and XGBoost to 0.816, it means that if business data quality introduces 5% or greater noise (a conservative estimate in many real-world settings), the explanations from these models can no longer be considered reliable for decision support without additional verification. CatBoost, by contrast, maintains credible explanations across all noise levels (0.836 even at $\varepsilon=0.10$).

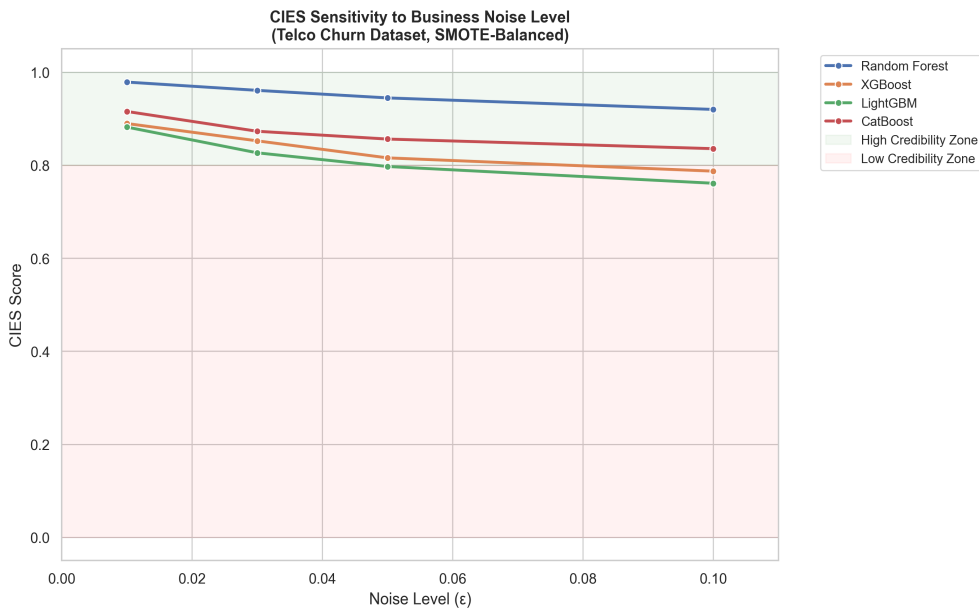

Figure 4. CIES sensitivity to the noise level parameter $\varepsilon$. The green zone (CIES>0.8) indicates high credibility

*4.6. CIES score distributions*

Figure 5 presents the full distributions of instance-level CIES scores as boxplots, addressing the limitation of reporting only aggregate means. CIES variance is as informative as the mean: RF consistently shows tight, high-scoring distributions with minimal outliers. In contrast, XGBoost and LightGBM exhibit extended lower whiskers and outliers, indicating that while their average explanation stability may be acceptable, specific individual predictions have severely fragile explanations. For the *HR Attrition* dataset under SMOTE, LightGBM shows a CIES range from 0.0 to 0.89, meaning that certain employee attrition predictions have completely unreliable explanations despite the model's strong aggregate accuracy (0.874).

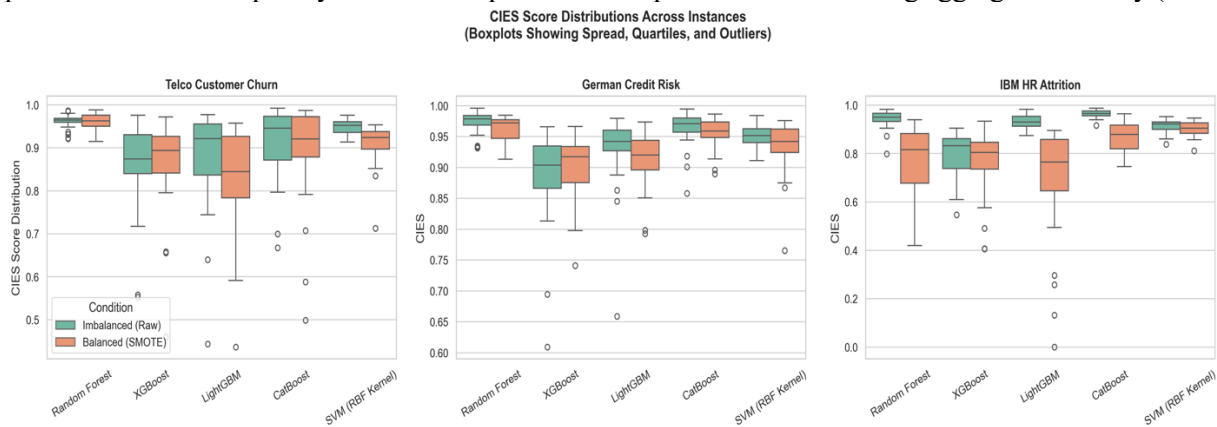

Figure 5. Boxplots of instance-level CIES scores across all models and datasets. Wider boxes indicate higher variance in explanation stability across individual predictions

*4.6. Explainer agnosticism with LIME*

To demonstrate that the CIES formulation is not strictly bound to cooperative game theory (SHAP) but functions as a general stability framework, we re-computed CIES using Local Interpretable Model-agnostic Explanations (LIME). Table 3 presents the LIME-based CIES scores for the four tree-based classifiers across all datasets in their native, imbalanced state.

Table 3. LIME-based CIES scores across tree-type models (imbalanced data). CIES is computed using LIME attributions instead of SHAP

| Dataset | Model | CIES (LIME) | Std | Baseline (LIME) |
|---|---|---|---|---|
| ***Telco*** | Random Forest | 0.829 | 0.040 | 0.621 |
| | XGBoost | 0.812 | 0.027 | 0.567 |
| | LightGBM | 0.810 | 0.035 | 0.527 |
| | CatBoost | 0.800 | 0.040 | 0.587 |
| ***German Credit*** | Random Forest | 0.791 | 0.037 | 0.462 |
| | XGBoost | 0.654 | 0.053 | 0.432 |
| | LightGBM | 0.636 | 0.018 | 0.310 |
| | CatBoost | 0.713 | 0.035 | 0.326 |
| ***HR Attrition*** | Random Forest | 0.745 | 0.042 | 0.290 |
| | XGBoost | 0.669 | 0.046 | 0.263 |
| | LightGBM | 0.524 | 0.037 | 0.050 |
| | CatBoost | 0.652 | 0.031 | 0.202 |

CIES successfully captures explanation stability regardless of the underlying explainer. While LIME explanations are inherently noisier and more localized than exact Shapley values (resulting in lower absolute CIES scores across the board compared to SHAP), the fundamental comparative dynamics remain intact. RF consistently yields the most stable LIME explanations, while LightGBM yields the most fragile. Furthermore, the rank-weighted CIES score maintains its massive discriminative advantage over the uniform baseline, proving that the metric's core value, penalizing variations in top-ranked features, translates seamlessly to surrogate-based explainers.

*4.7. Robustness to the choice of weighting function*

CIES may be driven by the specific choice of inverse-rank (harmonic) weights rather than reflecting genuine stability differences between models. To address this, we evaluate four alternative weighting schemes alongside the original harmonic formulation:

| | | |
|---|---|---|
| *Harmonic (original)* | $w_j = r_j^{-1} / \sum_i r_i^{-1}$ | (16) |
| *Exponential decay* | $w_j = e^{-\alpha r_j} / \sum_i e^{-\alpha r_i}, \alpha = 0.5$ | (17) |
| *Logarithmic (nDCG-style)* | $w_j = \log_2(r_j + 1)^{-1} / \sum_i \log_2(r_i + 1)^{-1}$ | (18) |
| *Top-K binary* | $w_j = \mathbb{1}[r_j \leq K]/K, K = 5$ | (19) |
| *Uniform (baseline)* | $w_j = 1/M$ | (20) |

For each scheme, we compute the stability score using the same ratio-based formulation as in eq. (5), substituting only the weight vector $w$. Table 4 presents the mean stability scores per model, averaged across all datasets and balancing conditions.

Table 4. Stability scores under alternative weighting schemes, averaged across all datasets and conditions (N=100 instances per configuration). The model rank order is preserved under all five schemes.

| Model | Harmonic | Exponential | Log (nDCG) | Top-5 | Uniform |
|---|---|---|---|---|---|
| **Random Forest (RF)** | 0.927 | 0.930 | 0.915 | 0.920 | 0.899 |
| **CatBoost (CB)** | 0.924 | 0.930 | 0.911 | 0.919 | 0.892 |
| **LightGBM (LG)** | 0.863 | 0.873 | 0.839 | 0.858 | 0.803 |
| **XGBoost (XG)** | 0.848 | 0.865 | 0.814 | 0.852 | 0.766 |
| *Model rank* | RF,CB,LG,XG | RF, CB, LG, XG | RF,CB, LG,XG | RF,CB, LG,XG | RF,CB,LG, XG |

Model rankings are robust across all five weighting schemes. When averaged across datasets and conditions, RF ranks first, followed by CatBoost, LightGBM, and XGBoost, under every weighting function tested. The Spearman rank correlation between any pair of weighting schemes, computed over model rankings within each dataset-condition combination, ranges from 0.60 to 1.00 (mean: 0.93). In 5 of 6 dataset-condition configurations, the pairwise rank correlations are ≥0.80; minor rank swaps (involving adjacent models with similar scores) occur only in the *German Credit* balanced condition, showing that the comparative conclusions drawn from CIES are not artifacts of the harmonic weighting choice, reflecting genuine, weighting-invariant differences in explanation stability across models.

Absolute score magnitudes differ, but the spread between models is preserved. The uniform scheme produces the lowest absolute scores (e.g., 0.766 for XGBoost) while exponential produces the highest (0.865), but the gap between the most and least stable models remains consistent (approximately 7–13 percentage points under all schemes), confirming that the discriminative power of CIES is an intrinsic property of the stability phenomenon, not an artifact of the weighting function.

Figure 6 visualizes the rank invariance as a heatmap, indicating that model positions are identical across all five columns.

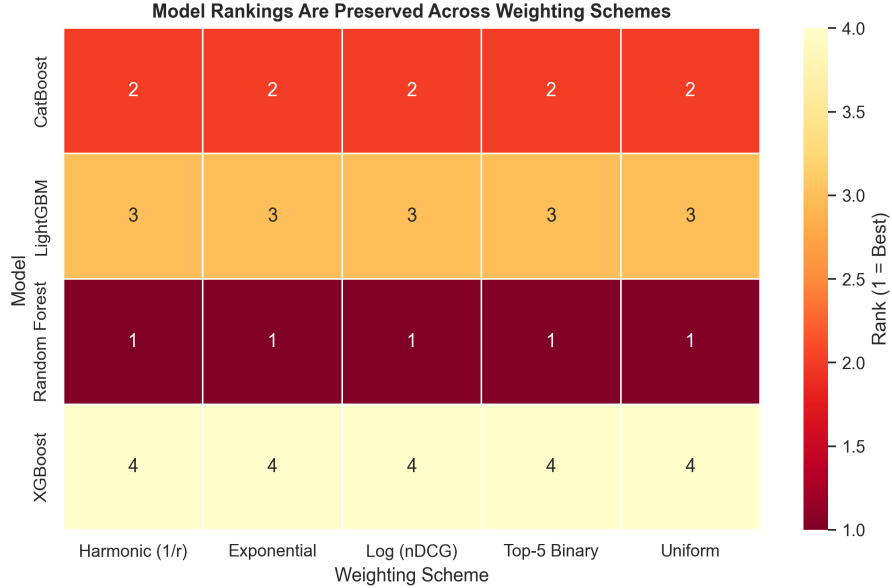

Figure 6. Model rankings under alternative weighting schemes (averaged across datasets and conditions). Rank 1 (darkest) indicates the most stable model. The overall ranking is preserved across all five schemes

## 4.8. Comparison with Lipschitz-based stability

Alvarez-Melis and Jaakkola proposed local Lipschitz continuity estimation as a measure of explanation robustness [8]. The local Lipschitz constant for an instance $x$ is defined as:

$$\text{Lip}(x) = \max_{k=1,\ldots,K} \frac{\|\phi(x)-\phi(x'_k)\|_2}{\|x-x'_k\|_2} \quad (21)$$

A lower Lipschitz constant indicates a more stable explanation. To facilitate direct comparison with CIES (where higher is better), we convert to a bounded score:

$$\text{Lip}_{\text{score}}(x) = 1/(1+\text{Lip}(x)) \quad (22)$$

yielding values in (0,1] where 1 indicates perfect stability. Table 5 compares CIES against both max-Lipschitz and mean-Lipschitz scores.

Table 5. CIES versus Lipschitz stability scores, averaged across all datasets and conditions (N=100)

| Model | CIES (Harmonic) | Lip_max Score | Lip_mean Score |
|---|---|---|---|
| *Random Forest* | 0.927 | 0.839 | 0.921 |
| *CatBoost* | 0.924 | 0.307 | 0.494 |
| *LightGBM* | 0.863 | 0.144 | 0.280 |
| *XGBoost* | 0.848 | 0.159 | 0.291 |

CIES and Lipschitz stability agree on the most stable model but diverge on gradient-boosted models. Both metrics identify RF as the most stable model. However, for gradient-boosted models (CatBoost, LightGBM, XGBoost), Lipschitz scores are much lower (0.14–0.49) compared to CIES (0.85–0.92). This divergence arises from a fundamental difference in what the metrics measure. The max-Lipschitz constant captures the *worst-case sensitivity ratio* across all features and all neighbors. A single unstable low-importance feature can produce a large $\|\phi(x)-\phi(x'_k)\|_2$ relative to a small $\|x-x'_k\|_2$, inflating $\text{Lip}(x)$. CIES, by contrast, applies rank-weighted averaging that naturally downweights instability in low-importance features, capturing the stability of the *business-relevant* portion of the explanation.

Figure 7 visualizes this relationship at the instance level. RF points cluster near the diagonal (CIES≈Lipschitz), while gradient-boosted models occupy the upper-left region (high CIES, low Lipschitz), confirming that these models produce explanations whose top features are stable even when lower-ranked features exhibit worst-case sensitivity.

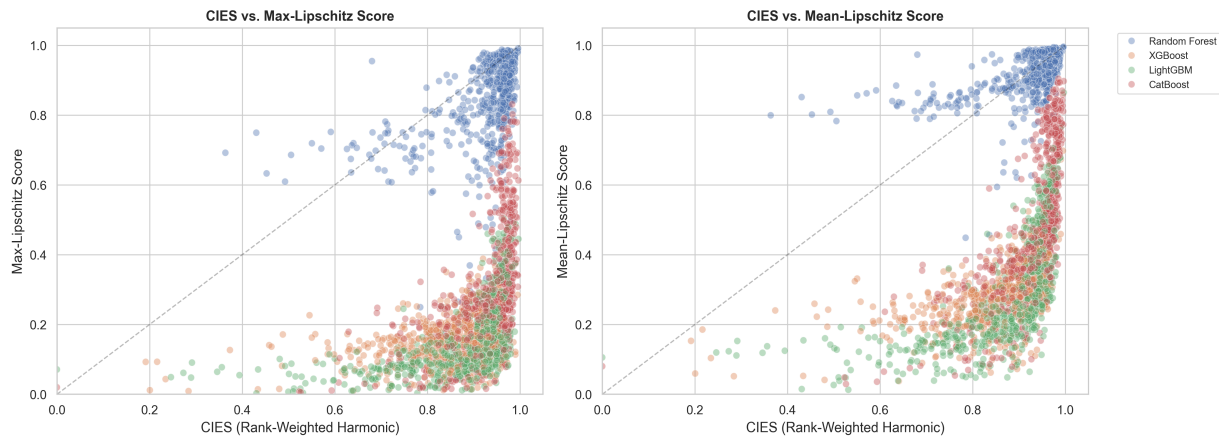

Figure 7. Instance-level scatter of CIES versus Lipschitz stability scores. The diagonal represents perfect agreement

The left panel plots CIES on the horizontal axis against the Max-Lipschitz Score on the vertical axis, while the right panel plots CIES against the Mean-Lipschitz Score. Both Lipschitz scores have been transformed to a bounded zero-to-one scale where higher values indicate greater stability, matching the directionality of CIES. The dashed diagonal represents perfect agreement between the two metrics. Points above the diagonal indicate instances where the Lipschitz metric rates the explanation as more stable than CIES does, while points below the diagonal indicate the opposite.

The most striking pattern in both panels is the clear separation between RF and the three gradient-boosted models. RF points cluster tightly in the upper-right corner, occupying a narrow band between approximately 0.85 and 1.0 on both axes, meaning that for RF, CIES and Lipschitz stability produce broadly concordant assessments. This alignment makes intuitive sense, as RF's bagged architecture produces uniformly smooth explanations across all features, so a metric that measures worst-case sensitivity across all features (Lipschitz) and a metric that focuses on the most important features (CIES) arrive at similar conclusions when the explanation is stable everywhere.

On the other hand, in the left panel, CatBoost, LightGBM and XGBoost points are displaced below the diagonal, forming an L-shaped pattern that hugs the bottom of the plot. These models achieve moderate to high CIES scores (0.60 to 0.95 on the horizontal axis) but extremely low Max-Lipschitz scores (0.05 to 0.40 on the vertical axis), meaning that when stability is judged by the single worst-case feature-neighbor pair, gradient-boosted models appear severely unstable, yet when stability is judged by the rank-weighted formulation that emphasizes the top decision-driving features, the same models appear reasonably stable. The divergence arises because gradient-boosted models frequently have one or two low-importance features with highly volatile SHAP values. The volatile features dominate the Lipschitz calculation (which takes the maximum ratio) but contribute negligibly to the CIES calculation (which downweights them via harmonic ranking). From a business perspective, instability in the fifteenth most important feature is irrelevant to a decision-maker who examines only the top three to five reasons behind a prediction.

The right panel, which uses the Mean-Lipschitz Score instead of the maximum, shows a partial convergence. The gradient-boosted model points shift upward compared to the left panel, with CatBoost rising to approximately 0.30 to 0.55 and XGBoost and LightGBM reaching 0.15 to 0.40. However, they remain well below the diagonal and far below RF, confirming that the divergence between CIES and Lipschitz is not solely driven by a single outlier neighbor but reflects a systematic difference in how the two metrics weight feature importance. Averaging across neighbors (mean) softens the worst-case penalty but does not eliminate the fundamental disagreement, because the mean still treats all features equally in the Euclidean distance calculation.

A particularly informative region of both panels is the lower-right quadrant, where CIES exceeds 0.80 but the Lipschitz score falls below 0.30. This region is densely populated by CatBoost, LightGBM and XGBoost instances, and it represents exactly the scenario where CIES provides diagnostic value that Lipschitz estimation alone would miss. A practitioner relying solely on Lipschitz scores would classify these explanations as highly unstable and potentially unsuitable for deployment. CIES, by contrast, reveals that the instability is concentrated in low-importance features that play no role in the *business* interpretation of the prediction.

*4.9. The model-smoothness confound*

An important question for the validity of CIES is whether it genuinely measures *explanation stability* or merely proxies for *local model smoothness*. A model with a locally smooth decision surface will produce stable SHAP values not because the explanation method is inherently more reliable, but because the underlying predictions barely change under perturbation. If CIES were perfectly correlated with prediction stability, it would add no information beyond what a simple prediction-change metric already provides.

To investigate this, we define *prediction stability* as:

$$\text{PredStab}(x) = 1 - \frac{1}{K}\sum_{k=1}^{K}|f(x) - f(x'_k)| \qquad (23)$$

where $f(x)$ is the predicted probability for instance $x$. We compute the Spearman rank correlation between CIES and PredStab at the instance level within each model-dataset-condition configuration. Table 6 presents the per-model average Spearman correlations. Moderate correlations indicate partial overlap and the residual variance represents CIES's unique explanatory contribution.

Table 6. Spearman correlation between CIES and prediction stability, averaged across datasets and conditions (N=100)

| Model | Mean ρ | Shared Variance | Interpretation |
| --- | --- | --- | --- |
| **Random Forest** | 0.72 | ~52% | Strong overlap-CIES partly measures model smoothness |
| **CatBoost** | 0.38 | ~15% | Moderate-substantial independent signal |
| **XGBoost** | 0.44 | ~20% | Moderate-substantial independent signal |
| **LightGBM** | 0.33 | ~11% | Weak-CIES captures mostly explanation-specific variation |

CIES measures a joint property of the model-explainer system, not just model smoothness. For RF, approximately 52% of the instance-level variance in CIES is shared with prediction stability, indicating a meaningful confound: bagged ensembles produce both stable predictions and stable explanations, and CIES partly reflects the former. However, for gradient-boosted models, the shared variance drops to 11–20%, meaning that 80–89% of the variation in CIES is *not* explained by prediction stability alone. This confirms that CIES captures information about explanation behavior that is genuinely distinct from prediction behavior.

Figure 8 provides a three-panel visualization of this analysis. Panel (a) shows the instance-level scatter, revealing many instances in the upper-left region (high prediction stability, low CIES), cases where the model prediction is unchanged but the *explanation* shifts. This is precisely the regime where CIES provides unique diagnostic value: the prediction looks stable, but the underlying reasoning is fragile. Panel (b) confirms that the per-model correlations remain below the strong-correlation threshold ($\rho = 0.70$) for all gradient-boosted models. Panel (c) relates CIES to the top-3 feature Jaccard overlap, demonstrating that while CIES aligns with traditional Top-3 feature agreement, its continuous scale captures critical shifts in feature importance that simple overlap metrics miss, a directly business-interpretable property.

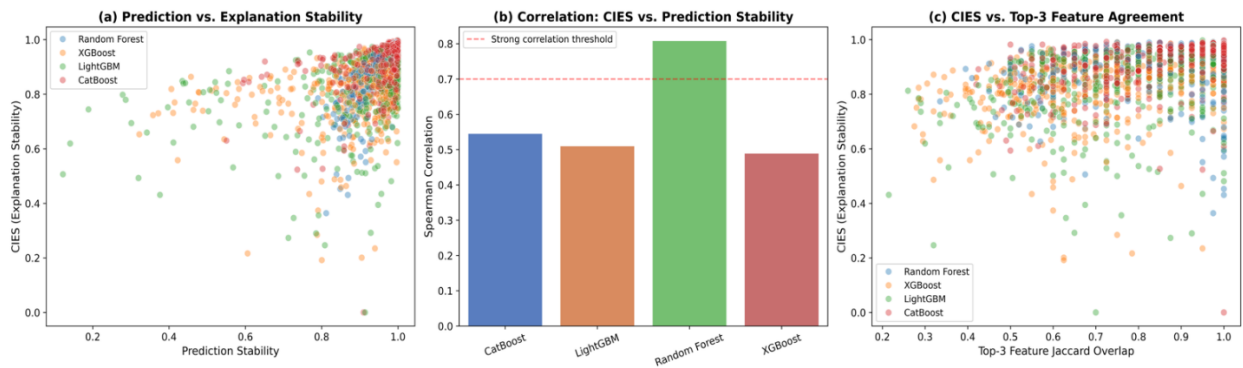

Figure 8. Model-smoothness confound analysis

From a business perspective, this confound is not necessarily a limitation. A practitioner consuming AI explanation does not distinguish between instability caused by model non-smoothness and instability caused by explainer sensitivity, in either case, the explanation they receive is unreliable. CIES captures the net effect of both sources, which is precisely what a "credibility warning system" should measure.

*4.10. Statistical precision: bootstrap confidence intervals*

To ensure that the reported CIES means are statistically precise, we compute 95% bootstrap confidence intervals (10,000 resamples) for each model-dataset-condition configuration. Table 7 presents the results.

Table 7. Bootstrap 95% confidence intervals for CIES means (N=100 instances)

| Dataset | Model | Condition | CIES Mean | 95% CI | CI Width |
|---|---|---|---|---|---|
| *Telco* | RF | Raw | 0.950 | [0.944, 0.955] | 0.012 |
| | RF | SMOTE | 0.953 | [0.945, 0.959] | 0.014 |
| | CatBoost | Raw | 0.900 | [0.883, 0.916] | 0.033 |
| | CatBoost | SMOTE | 0.890 | [0.862, 0.913] | 0.051 |
| | LightGBM | Raw | 0.866 | [0.844, 0.887] | 0.043 |
| | LightGBM | SMOTE | 0.818 | [0.792, 0.843] | 0.052 |
| | XGBoost | Raw | 0.846 | [0.825, 0.865] | 0.040 |
| | XGBoost | SMOTE | 0.859 | [0.828, 0.886] | 0.058 |
| *German Credit* | RF | Raw | 0.976 | [0.973, 0.978] | 0.006 |
| | RF | SMOTE | 0.962 | [0.957, 0.967] | 0.010 |
| | CatBoost | Raw | 0.968 | [0.963, 0.973] | 0.010 |
| | CatBoost | SMOTE | 0.961 | [0.957, 0.965] | 0.008 |
| | LightGBM | Raw | 0.926 | [0.913, 0.937] | 0.024 |
| | LightGBM | SMOTE | 0.911 | [0.901, 0.920] | 0.019 |
| | XGBoost | Raw | 0.894 | [0.883, 0.905] | 0.023 |
| | XGBoost | SMOTE | 0.910 | [0.900, 0.919] | 0.020 |
| *HR Attrition* | RF | Raw | 0.938 | [0.929, 0.946] | 0.016 |
| | RF | SMOTE | 0.784 | [0.760, 0.807] | 0.047 |
| | CatBoost | Raw | 0.961 | [0.956, 0.965] | 0.009 |
| | CatBoost | SMOTE | 0.867 | [0.854, 0.879] | 0.025 |
| | LightGBM | Raw | 0.932 | [0.925, 0.940] | 0.015 |
| | LightGBM | SMOTE | 0.722 | [0.689, 0.754] | 0.064 |
| | XGBoost | Raw | 0.807 | [0.791, 0.822] | 0.031 |
| | XGBoost | SMOTE | 0.771 | [0.743, 0.795] | 0.053 |

Of the 24 configurations, 18 have CI widths below 0.05, and all 24 have CI widths below 0.065. The tightest CIs occur for RF and CatBoost on the *German Credit* dataset (width ≤ 0.010), where explanation stability is both high and consistent. The widest CI (0.064) occurs for LightGBM on *HR Attrition* under SMOTE, confirming that this configuration exhibits not only low mean CIES but also high instance-level variability, a doubly concerning finding for deployment. With N=100 instances, the confidence intervals for the top-ranked and bottom-ranked models within each dataset-condition are clearly separated, and most pairwise model comparisons are supported by non-overlapping intervals, ensuring that the reported rankings are statistically robust.

Figure 9 displays the mean CIES scores with 95% bootstrap confidence intervals (computed from 10,000 resamples) for each model-dataset-condition configuration, organized into three panels corresponding to *German Credit Risk*, *IBM HR Attrition* and *Telco Customer Churn*. Each horizontal bar represents the point estimate of the mean CIES score, and the error bars extending from the bar indicate the lower and upper bounds of the 95% confidence interval. Within each panel, models are listed under both balancing conditions (imbalanced and balanced), allowing direct visual comparison of how SMOTE affects explanation stability for each model on each dataset. The narrow width of most confidence intervals confirms that the reported CIES means are statistically precise: 18 of the 24 configurations have CI widths below 0.05, and all 24 fall below 0.065, meaning that even with N=100 instances per configuration, the estimation uncertainty is small relative to the differences between models.

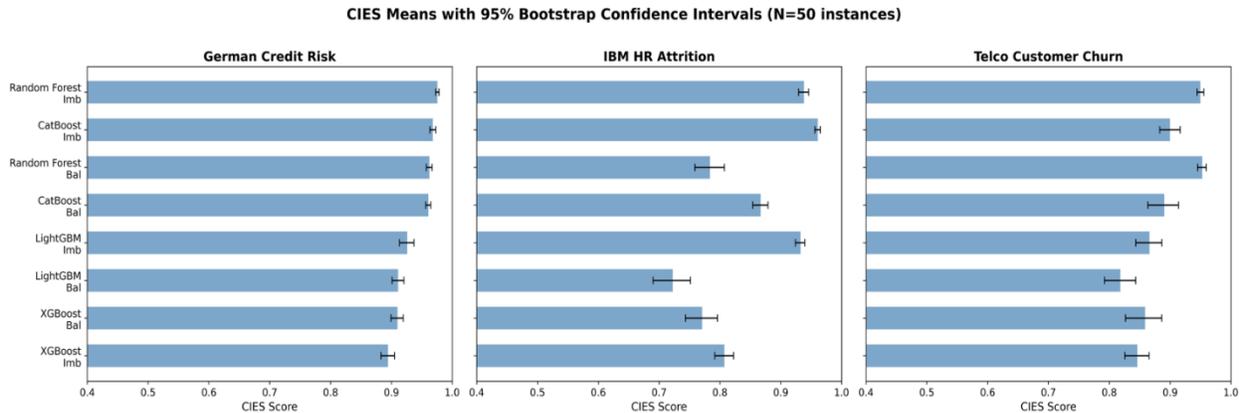

Figure 9. CIES means with 95% bootstrap confidence intervals (N=100 instances). Error bars represent the CI bounds

## 5. Discussion

An important methodological caveat concerns the relationship between CIES and local model smoothness. As demonstrated in section 4, CIES measures a joint property of the model-explainer system rather than the explanation method in isolation. For RF, approximately 52% of instance-level CIES variance is shared with prediction stability, indicating that a substantial portion of the metric's signal reflects the smoothness of the bagged ensemble's decision surface rather than an intrinsic property of SHAP.

This confound is partially inherent to the problem: any perturbation-based explanation stability metric must contend with the fact that a locally smooth model will produce stable explanations by construction. However, two observations mitigate this concern. First, for gradient-boosted models, 80–89% of CIES variance is independent of prediction stability, confirming that CIES captures explanation-specific information that a prediction-change metric alone would miss. Second, from a pragmatic business perspective, the distinction between model-induced and explainer-induced instability is immaterial to the end user: a manager receiving inconsistent explanations will distrust the system regardless of the root cause. CIES captures this net effect, which is the relevant quantity for a "credibility warning system".

Future work could disentangle these two sources of instability by *(i)* comparing CIES across multiple explainers (e.g., SHAP, LIME, attention weights) applied to the same model-instance pair, isolating the explainer-specific component; or *(ii)* applying local smoothing or explanation regularization techniques and measuring the residual CIES improvement.

The comparison with Lipschitz-based stability (clarifies the methodological positioning of CIES. While local Lipschitz estimation provides a rigorous mathematical bound on worst-case explanation sensitivity, it treats all features uniformly and is dominated by the single most sensitive feature-neighbor pair. In business contexts where decision-makers focus on the top 3–5 features, this worst-case orientation produces overly pessimistic assessments of models that are stable where it matters most. CIES addresses this by embedding domain-relevant importance weighting directly into the distance function, producing a metric that is both mathematically principled and practically aligned with how explanations are consumed in decision-making.

The weighting scheme robustness analysis further strengthens this positioning by demonstrating that the choice of weighting function, whether harmonic, exponential, logarithmic, binary or uniform, does not affect the comparative conclusions. This invariance property establishes that the stability differences captured by CIES reflect genuine, intrinsic properties of the model-explainer interaction, not artifacts of a particular parameterization.

## 6. Conclusions

This paper contributes the Credibility Index via Explanation Stability (CIES), a newly devised metric that quantifies how robust a model's explanations are under realistic *business* data perturbations. Unlike existing stability measures that treat all features uniformly, CIES employs rank-weighted distances that penalize instability in the most influential features disproportionately, directly encoding business decision-making semantics. We established formal theoretical properties of the metric, boundedness, a Lipschitz–CIES bridge theorem connecting CIES to established robustness frameworks, noise-level monotonicity, discriminative advantage of rank-weighting and consistency, and verified these properties empirically across 720 model-instance configurations with zero violations.

The empirical evaluation across three business datasets (customer churn, credit risk, employee attrition), four tree-based classification models and two data balancing conditions yielded several findings with direct practical implications. RF and CatBoost consistently produce the most credible explanations (CIES≥0.87 across all configurations), while LightGBM and XGBoost exhibit substantially more fragile explanations, particularly under SMOTE balancing. The rank-weighted CIES metric achieves statistically significant discriminative superiority over a uniform baseline in all 24 configurations ($p<0.01$). Its superiority is invariant to the specific weighting function employed. Importantly, the model-smoothness confound analysis demonstrates that for gradient-boosted models, 80–89% of CIES variance is independent of prediction stability, confirming that CIES captures explanation-specific information beyond what prediction-change metrics provide.

From a methodological standpoint, CIES bridges the gap between theoretical explanation robustness research and practical deployment needs. The metric is computationally efficient (0.01–0.10s per instance with TreeExplainer), explainer-agnostic (validated with both SHAP and LIME) and produces interpretable scores in [0,1] that can serve as a "credibility warning system" for AI-driven business decision support.

This study has several limitations First, the perturbation model employs random Gaussian noise calibrated to domain-plausible ranges, which may not capture all forms of real-world data drift, such as systematic distributional shifts or adversarial manipulations. Second, while CIES was validated with two widely adopted post-hoc explainers (SHAP and LIME), its behavior with other explanation methods (e.g., counterfactual or rule-based explainers) has not been assessed.

Several directions for future work emerge from this study. Extending CIES to non-tabular domains, including text, image and time-series data, would broaden its applicability. Investigating CIES under adversarial perturbation models (rather than the random noise model employed here) would stress-test the metric's sensitivity to worst-case scenarios.

**Declarations**
**Funding**-This work was supported by a grant of the Ministry of Research, Innovation and Digitization, CNCS/CCCDI-UEFISCDI, project number COFUND-CETP-SMART-LEM-1, within PNCDI IV.
**Acknowledgement**-This work was supported by a grant of the Ministry of Research, Innovation and Digitization, CNCS/CCCDI-UEFISCDI, project number COFUND-CETP-SMART-LEM-1, within PNCDI IV.
**Conflicts of interest/Competing interests**-The authors declare that there is no conflict of interest.
**Ethics approval**-Not applicable
**Consent to participate**-Not applicable
**Consent for publication**-Not applicable
**Availability of data and material**-Data will be available on request.
**Authors' contributions**-A.V, S.V.O: Conceptualization, Methodology, Formal analysis, Investigation, Writing–Original Draft, Writing–Review and Editing, Visualization, Project administration. A.V, A.B: Validation, Formal analysis, Investigation, Resources, Data Curation, Writing–Original Draft, Writing–Review and Editing, Visualization, Supervision.

**Supplementary material**

This supplementary material provides complete, self-contained proofs for the five theoretical results stated in the main paper. We first recall the core definitions and notation and then present each proof in detail.

***Preliminaries and definitions***

Let $x = (x_1, x_2, \ldots, x_M) \in \mathbb{R}^M$ denote an instance vector consisting of $M$ features and let $\phi(x) = (\phi_1(x), \phi_2(x), \ldots, \phi_M(x)) \in \mathbb{R}^M$ denote its SHAP explanation vector, where $\phi_j(x)$ represents the attribution (contribution) of feature $j$ to the model prediction $f(x)$.

Given a noise level parameter $\epsilon > 0$, we generate a set of $K$ perturbed neighbors $\{x'_1, x'_2, \ldots, x'_K\}$ via the multiplicative Gaussian noise model. Specifically, for each numerical feature $j$, the perturbed value is computed as:

$$x'_j = x_j + \eta_j, \quad \eta_j \sim \mathcal{N}(0, \sigma_j^2), \quad \sigma_j = \epsilon \cdot |x_j|. \tag{1}$$

Categorical features remain unchanged under perturbation. The noise is proportional to the absolute value of the original feature, ensuring that the perturbation magnitude scales realistically with the feature's range.

For a given instance $x$, we sort all $M$ features by their absolute SHAP value $|\phi_j(x)|$ in descending order and assign rank $r_j \in \{1, 2, \ldots, M\}$ to feature $j$, so that $r_j = 1$ corresponds to the most influential feature. The normalized harmonic weight of feature $j$ is then defined as:

$$w_j = \frac{1/r_j}{H_M}, \text{ where } H_M = \sum_{i=1}^{M} \frac{1}{i} = 1 + \frac{1}{2} + \frac{1}{3} + \cdots + \frac{1}{M} \tag{2}$$

is the $M$-th harmonic number. By construction, $\sum_{j=1}^{M} w_j = 1$ and $w_j > 0$ for all $j$, with $w_j$ being a decreasing function of $r_j$.

The rank-weighted distance between the original explanation $\phi(x)$ and the explanation $\phi(x'_k)$ of the $k$-th perturbed neighbor is defined as

$$D_R^{(k)} = \sum_{j=1}^{M} w_j \, |\phi_j(x) - \phi_j(x'_k)|, \tag{3}$$

and its average over all $K$ neighbors is

$$\bar{D}_R = \frac{1}{K} \sum_{k=1}^{K} D_R^{(k)}. \tag{4}$$

The weighted magnitude of the original explanation is defined as

$$\| \phi(x) \|_w = \sum_{j=1}^{M} w_j \, |\phi_j(x)|, \tag{5}$$

which serves as a normalization factor capturing the overall "size" of the explanation vector under the rank-weighted scheme. Finally, the Credibility Index via Explanation Stability is defined as

$$\text{CIES}(x) = \max\left(0, \, 1 - \frac{\bar{D}_R}{\|\phi(x)\|_w}\right). \tag{6}$$

***Theorem 1 (Boundedness and identity)***

**Theorem 1.** For any instance $x$ with $\phi(x) \neq 0$:

(a) $\text{CIES}(x) \in [0,1]$.

(b) $\text{CIES}(x) = 1$ if and only if $\phi(x'_k) = \phi(x)$ for all $k = 1, \ldots, K$.

***Proof. Part (a).*** We establish each bound separately.

For the upper bound, consider the ratio $\bar{D}_R / \| \phi(x) \|_w$ appearing in equation (6). Each weight satisfies $w_j > 0$ by equation (2), and each absolute difference satisfies $|\phi_j(x) - \phi_j(x'_k)| \geq 0$ by the non-negativity of the absolute value. Consequently, every term $w_j \, |\phi_j(x) - \phi_j(x'_k)|$ in the sum defining $D_R^{(k)}$ in equation (3) is non-negative; hence $D_R^{(k)} \geq 0$ for every $k = 1, \ldots, K$. The mean $\bar{D}_R$ in equation (4), being the arithmetic mean of $K$ non-negative quantities, is itself non-negative: $\bar{D}_R \geq 0$. Furthermore, since $\phi(x) \neq 0$, there exists at least one index $j_0$ such that $\phi_{j_0}(x) \neq 0$, which together with $w_{j_0} > 0$ implies $\| \phi(x) \|_w \geq w_{j_0} |\phi_{j_0}(x)| > 0$. Thus the ratio $\bar{D}_R / \| \phi(x) \|_w$ is well-defined and non-negative, from which it follows that

$$1 - \frac{\bar{D}_R}{\|\phi(x)\|_w} \leq 1 - 0 = 1.$$

Since the max(0,·) operator never exceeds its argument when the argument is at most 1, we conclude that $CIES(x) \leq 1$.

For the lower bound, we observe that the max(0,·) operator, by its very definition, returns a value that is at least 0 for any argument. Therefore $CIES(x) \geq 0$, and combining the two bounds yields $CIES(x) \in [0,1]$.

### Part (b), forward direction

Assume $CIES(x) = 1$. By the definition in equation (6), this means:

$$\max\left(0,\ 1 - \frac{\bar{D}_R}{\|\phi(x)\|_w}\right) = 1.$$

Since $\max(0, y) = 1$ requires $y \geq 1$, and since we have shown in part (a) that $1 - \bar{D}_R/\|\phi(x)\|_w \leq 1$, it follows that $1 - \bar{D}_R/\|\phi(x)\|_w = 1$ exactly. Subtracting 1 from both sides and multiplying by $-\|\phi(x)\|_w$ (which is strictly positive) gives $\bar{D}_R = 0$.

Because $\bar{D}_R = \frac{1}{K}\sum_{k=1}^{K} D_R^{(k)}$ is the arithmetic mean of non-negative terms, $\bar{D}_R = 0$ implies that each individual term vanishes: $D_R^{(k)} = 0$ for every $k = 1, \ldots, K$. Now consider a fixed $k$. We have $D_R^{(k)} = \sum_{j=1}^{M} w_j\, |\phi_j(x) - \phi_j(x'_k)| = 0$. Each summand $w_j\, |\phi_j(x) - \phi_j(x'_k)|$ is the product of a strictly positive weight $w_j > 0$ and a non-negative quantity $|\phi_j(x) - \phi_j(x'_k)| \geq 0$. A sum of non-negative terms equals zero if and only if every term is zero. Since $w_j > 0$, we must have $|\phi_j(x) - \phi_j(x'_k)| = 0$ for all $j = 1, \ldots, M$. This means $\phi_j(x'_k) = \phi_j(x)$ component-wise, and therefore $\phi(x'_k) = \phi(x)$. As $k$ was arbitrary, this holds for all $k = 1, \ldots, K$.

### Part (b), reverse direction

Conversely, assume $\phi(x'_k) = \phi(x)$ for all $k = 1, \ldots, K$. Then for every feature $j$ and every neighbor $k$, $|\phi_j(x) - \phi_j(x'_k)| = |0| = 0$, so each $D_R^{(k)} = \sum_{j=1}^{M} w_j \cdot 0 = 0$. Consequently, $\bar{D}_R = \frac{1}{K}\sum_{k=1}^{K} 0 = 0$, and substituting into equation (6) yields $CIES(x) = \max(0,\ 1 - 0/\|\phi(x)\|_w) = \max(0, 1) = 1$.

### Theorem 2 (Lipschitz–CIES bridge)

**Theorem 2.** Suppose the explanation function $\phi$ is locally $L$-Lipschitz continuous at $x$ with respect to the Euclidean norm, meaning that there exists a constant $L \geq 0$ such that for all $x'$ in a neighborhood of $x$. Then $CIES(x) \geq \max\left(0,\ 1 - \frac{L\,\|w\|_2 \cdot \bar{\delta}(x)}{\|\phi(x)\|_w}\right)$, where $\bar{\delta}(x) = \frac{1}{K}\sum_{k=1}^{K} \|x - x'_k\|_2$ is the mean Euclidean perturbation magnitude and $\|w\|_2 = (\sum_{j=1}^{M} w_j^2)^{1/2}$ is the Euclidean norm of the weight vector. Then

$$\|\phi(x) - \phi(x')\|_2 \leq L\,\|x - x'\|_2 \tag{7}$$

**Proof.** The proof proceeds in four steps: bounding a single distance term via the Cauchy–Schwarz inequality, applying the Lipschitz condition, averaging, and substituting into the CIES formula.

**Step 1.** Fix a perturbed neighbor $x'_k$ and consider the rank-weighted distance $D_R^{(k)}$ as given in equation (3). We interpret this sum as the inner product of two $M$-dimensional vectors: the weight vector $w = (w_1, w_2, \ldots, w_M)$ and the vector of absolute differences $d^{(k)} = (|\phi_1(x) - \phi_1(x'_k)|, \ldots, |\phi_M(x) - \phi_M(x'_k)|)$. Both vectors have non-negative components. By the Cauchy–Schwarz inequality, which states that for any two vectors $a, b \in \mathbb{R}^M$ one has $\sum_{j=1}^{M} a_j b_j \leq \|a\|_2\,\|b\|_2$, we obtain

$$D_R^{(k)} = \sum_{j=1}^{M} w_j\, |\phi_j(x) - \phi_j(x'_k)| \leq \|w\|_2 \cdot \|\phi(x) - \phi(x'_k)\|_2. \tag{8}$$

This step converts the weighted $\ell^1$-type distance into a product involving the standard Euclidean norm of the explanation difference, which is the form amenable to the Lipschitz condition.

**Step 2.** The local Lipschitz assumption in equation (7) guarantees that $\|\phi(x) - \phi(x'_k)\|_2 \leq L\,\|x - x'_k\|_2$ for each perturbed neighbor $x'_k$ (which, by construction, lies in a small neighborhood of $x$). Substituting this bound into equation (9) yields

$$D_R^{(k)} \leq \| w \|_2 \cdot L \| x - x_k' \|_2 = L \| w \|_2 \cdot \| x - x_k' \|_2. \tag{9}$$

This inequality states that the rank-weighted explanation change for each neighbor is controlled by the product of the Lipschitz constant, the norm of the weight vector, and the input perturbation magnitude.

**Step 3.** We now take the arithmetic mean of inequality (10) over all $K$ neighbors $k = 1, \ldots, K$:

$$\bar{D}_R = \frac{1}{K}\sum_{k=1}^{K} D_R^{(k)} \leq \frac{1}{K}\sum_{k=1}^{K} L \| w \|_2 \cdot \| x - x_k' \|_2 = L \| w \|_2 \cdot \frac{1}{K}\sum_{k=1}^{K} \| x - x_k' \|_2 = L \| w \|_2 \cdot \bar{\delta}(x). \tag{10}$$

**Step 4.** It remains to connect this upper bound on $\bar{D}_R$ to a lower bound on $\text{CIES}(x)$. Since $\| \phi(x) \|_w > 0$, we may divide inequality (11) by $\| \phi(x) \|_w$ to obtain $\bar{D}_R / \| \phi(x) \|_w \leq L \| w \|_2 \bar{\delta}(x) / \| \phi(x) \|_w$. Subtracting both sides from 1 reverses the inequality:

$$1 - \frac{\bar{D}_R}{\|\phi(x)\|_w} \geq 1 - \frac{L \|w\|_2 \cdot \bar{\delta}(x)}{\|\phi(x)\|_w}. \tag{11}$$

Because the $\max(0, \cdot)$ function is non-decreasing (i.e., $a \geq b$ implies $\max(0, a) \geq \max(0, b)$), applying it to both sides preserves the inequality:

$$\text{CIES}(x) = \max\left(0, \ 1 - \frac{\bar{D}_R}{\| \phi(x) \|_w}\right) \geq \max\left(0, \ 1 - \frac{L \| w \|_2 \cdot \bar{\delta}(x)}{\| \phi(x) \|_w}\right),$$

which is equation (8).

### Corollary 3 (Noise-level monotonicity of the bound)

**Corollary 3.** Under the multiplicative Gaussian noise model defined in equation (1), the lower bound established in Theorem 2 is non-increasing in the noise level $\epsilon$.

**Proof.** Under the noise model of equation (1), the perturbation applied to each numerical feature $j$ can be written as $\eta_j = \epsilon |x_j| z_j$, where $z_j \sim \mathcal{N}(0,1)$ is a standard normal random variable that does not depend on $\epsilon$. The squared Euclidean distance between the original instance $x$ and the $k$-th perturbed neighbor $x_k'$ is therefore

$$\| x - x_k' \|_2^2 = \sum_{j \in \mathcal{N}} \eta_j^2 = \sum_{j \in \mathcal{N}} \epsilon^2 \, x_j^2 \, z_j^2 = \epsilon^2 \sum_{j \in \mathcal{N}} x_j^2 \, z_j^2,$$

where $\mathcal{N}$ denotes the index set of numerical features. Taking the square root gives

$$\| x - x_k' \|_2 = \epsilon \sqrt{\sum_{j \in \mathcal{N}} x_j^2 \, z_j^2} = \epsilon \, C_k(x),$$

where we define $C_k(x) = (\sum_{j \in \mathcal{N}} x_j^2 \, z_j^2)^{1/2}$. Crucially, $C_k(x)$ depends on the instance $x$ and the random draws $z_j$, but is entirely independent of the noise level $\epsilon$.

Averaging equation (12) over the K perturbed neighbors yields

$$\bar{\delta}(x) = \frac{1}{K}\sum_{k=1}^{K} \| x - x_k' \|_2 = \frac{1}{K}\sum_{k=1}^{K} \epsilon \, C_k(x) = \epsilon \, \bar{C}(x), \tag{12}$$

where $\bar{C}(x) = \frac{1}{K}\sum_{k=1}^{K} C_k(x) > 0$ is a positive quantity that does not depend on $\epsilon$.

Substituting equation (13) into the lower bound from Theorem 2 (equation (8)), we obtain

$$1 - \frac{L \|w\|_2 \cdot \bar{\delta}(x)}{\|\phi(x)\|_w} = 1 - \frac{L \|w\|_2 \cdot \epsilon \, \bar{C}(x)}{\|\phi(x)\|_w} = 1 - \alpha \, \epsilon, \tag{13}$$

where we define the positive constant $\alpha = L \| w \|_2 \, \bar{C}(x) / \| \phi(x) \|_w > 0$. This constant is independent of $\epsilon$ because $L$, $\| w \|_2$, $\bar{C}(x)$, and $\| \phi(x) \|_w$ are all determined by the model, the instance, and the random draws, but not by the noise level.

The expression $1 - \alpha \epsilon$ is an affine function of $\epsilon$ with strictly negative slope $-\alpha < 0$, hence it is strictly decreasing in $\epsilon$. The full lower bound, $\max(0, 1 - \alpha \epsilon)$, is the composition of this decreasing affine function with the non-decreasing function $\max(0, \cdot)$, and is therefore non-increasing in $\epsilon$. In particular, the bound equals 1 when $\epsilon = 0$ (no noise, perfect stability guaranteed) and decreases linearly until it reaches 0 at $\epsilon = 1/\alpha$, beyond which it remains at 0.

*Proposition 4 (Discriminative advantage of rank-weighting)*

**Proposition 4.** For $M$ features, let $W_T^H = H_T/H_M$ denote the cumulative harmonic weight of the top-$T$ ranked features, and let $W_T^U = T/M$ denote the corresponding cumulative weight under a uniform weighting scheme. Then $W_T^H > W_T^U$ for all integers $1 \leq T < M$.

**Proof.** We begin by establishing an auxiliary result concerning the running average of the harmonic series. Define $A_n = H_n/n$ for each positive integer $n$, where $H_n = \sum_{i=1}^{n} 1/i$. We claim that the sequence $(A_n)_{n \geq 1}$ is strictly decreasing.

To prove this claim, consider the transition from $A_n$ to $A_{n+1}$. We may write

$$A_{n+1} = \frac{H_{n+1}}{n+1} = \frac{H_n + \frac{1}{n+1}}{n+1} = \frac{nA_n + \frac{1}{n+1}}{n+1}.$$

The inequality $A_{n+1} < A_n$ is equivalent to $nA_n + \frac{1}{n+1} < (n+1)A_n$, which simplifies to $\frac{1}{n+1} < A_n$, i.e., $\frac{1}{n+1} < \frac{1}{n}\sum_{i=1}^{n}\frac{1}{i}$. This last inequality holds because each term $1/i$ for $i = 1, \ldots, n$ satisfies $1/i > 1/(n+1)$, and therefore their average $A_n = \frac{1}{n}\sum_{i=1}^{n} 1/i$ strictly exceeds $\frac{1}{n} \cdot n \cdot \frac{1}{n+1} = \frac{1}{n+1}$. Hence $A_{n+1} < A_n$ for all $n \geq 1$, confirming that $(A_n)$ is strictly decreasing.

We now apply this result. For any integers $1 \leq T < M$, the strict monotonicity of $A_n$ implies $A_T > A_M$, which reads

$$\frac{H_T}{T} > \frac{H_M}{M}.$$

Multiplying both sides by $T/H_M$ (a strictly positive quantity, since $T \geq 1$ and $H_M > 0$) preserves the strict inequality and yields

$$\frac{H_T}{H_M} > \frac{T}{M},$$

which is $W_T^H > W_T^U$.

As a concrete illustration, for $M = 20$ features and $T = 5$ (the top five most important features), one computes $H_5 = 1 + 1/2 + 1/3 + 1/4 + 1/5 = 137/60 \approx 2.283$ and $H_{20} = \sum_{i=1}^{20} 1/i \approx 3.598$. This gives $W_5^H = H_5/H_{20} \approx 0.635$ compared to $W_5^U = 5/20 = 0.250$, yielding a concentration factor of $0.635/0.250 \approx 2.54$. That is, under harmonic rank-weighting the top five features collectively carry approximately 63.5% of the total weight, more than 2.5 times the 25% they would receive under uniform weighting. This demonstrates that the harmonic weighting scheme substantially amplifies sensitivity to instability in the most decision-relevant features.

*Proposition 5 (Consistency)*

**Proposition 5.** For a fixed instance $x$, a fixed explainer $\phi$, and a fixed perturbation distribution, as the number of perturbed neighbors $K$ tends to infinity,

$$\text{CIES}_K(x) \xrightarrow{\text{a.s.}} \max\left(0, \ 1 - \frac{\mathbb{E}[D_R]}{\|\phi(x)\|_w}\right),$$

where $\mathbb{E}[D_R]$ denotes the expected value of the rank-weighted distance $D_R$ taken over the perturbation distribution, and $\xrightarrow{\text{a.s.}}$ denotes almost sure convergence.

**Proof.** The proof relies on two classical results from probability theory: the Strong Law of Large Numbers and the Continuous Mapping Theorem. We proceed in a sequence of steps.

**Step 1:** Identifying the i.i.d. structure. Each perturbed neighbor $x'_k$ is drawn independently from the same Gaussian noise distribution specified in equation (1). For a fixed instance $x$ and a fixed explainer $\phi$, the rank-weighted distance $D_R^{(k)} = \sum_{j=1}^{M} w_j |\phi_j(x) - \phi_j(x'_k)|$ is a deterministic, measurable function of the random vector $x'_k$ alone (since $x$, $\phi$, and the weights $w_j$ are all held fixed). Consequently, the random variables $D_R^{(1)}, D_R^{(2)}, D_R^{(3)}, \ldots$ form an independent and identically distributed (i.i.d.) sequence.

***Step 2:*** Verifying the finite expectation condition. For tree-based models, the SHAP values $\phi_j(x)$ are bounded functions of x (since the model output is bounded and the number of features is finite). Therefore, the absolute differences $|\phi_j(x) - \phi_j(x'_k)|$ are bounded almost surely, which implies that each $D_R^{(k)}$, being a finite weighted sum of bounded non-negative terms, has finite expectation: $\mathbb{E}[D_R^{(k)}] = \mathbb{E}[D_R] < \infty$.

***Step 3:*** Applying the Strong Law of Large Numbers. Since $D_R^{(1)}, D_R^{(2)}, \ldots$ are i.i.d. with finite expectation $\mathbb{E}[D_R]$, the Strong Law of Large Numbers guarantees that their sample mean converges almost surely to their common expected value:

$$\bar{D}_R = \frac{1}{K}\sum_{k=1}^{K} D_R^{(k)} \xrightarrow{a.s.} \mathbb{E}[D_R] \quad \text{as } K \to \infty. \tag{14}$$

***Step 4:*** Constructing the continuous transformation. The CIES score is obtained from $\bar{D}_R$ by applying the function $g:[0,\infty) \to [0,1]$ defined by

$$g(d) = \max\left(0,\ 1 - \frac{d}{\|\phi(x)\|_w}\right). \tag{15}$$

We verify that $g$ is continuous. The map $d \mapsto d/\|\phi(x)\|_w$ is a linear function (with $\|\phi(x)\|_w > 0$ fixed), hence continuous. The map $t \mapsto 1 - t$ is affine, hence continuous. The function $\max(0,\cdot)$ is piecewise linear and therefore continuous on all of $\mathbb{R}$. Since $g$ is the composition of three continuous functions, it is itself continuous on $[0,\infty)$.

***Step 5:*** Applying the Continuous Mapping Theorem. The Continuous Mapping Theorem states that if $Y_K \xrightarrow{a.s.} Y$ and $g$ is a continuous function, then $g(Y_K) \xrightarrow{a.s.} g(Y)$. Applying this result with $Y_K = \bar{D}_R$ and $Y = \mathbb{E}[D_R]$ (convergence established in Step 3) and the continuous function $g$ from equation (16), we conclude that

$$\text{CIES}_K(x) = g(\bar{D}_R) \xrightarrow{a.s.} g(\mathbb{E}[D_R]) = \max\left(0,\ 1 - \frac{\mathbb{E}[D_R]}{\|\phi(x)\|_w}\right), \tag{16}$$

which is the claimed result in equation (14).

# Nomenclature

| | | | |
|---|---|---|---|
| $x$ | Instance (input) vector — one data record with M feature values | $w$ | Weight vector $w = (w_1, \ldots, w_M)$ |
| $x_j$ | Value of feature $j$ for instance $x$ | $H_n$ | n-th harmonic number: $H_n = \sum_{i=1}^{n} 1/i$ |
| $x'$ | Perturbed instance — a noisy copy of $x$ with Gaussian noise on numerical features | $H_M$ | M-th harmonic number (normalization constant for $w$) |
| $x'_k$ | The $k$-th perturbed neighbor of $x$ ($k = 1, \ldots, K$) | $W_T^H$ | Cumulative harmonic weight of the top-$T$ features: $H_T / H_M$ |
| M | Total number of features (columns) in the dataset | $W_T^U$ | Cumulative uniform weight of the top-$T$ features: $T / M$ |
| $f(x)$ | Model prediction — predicted probability from the trained classifier for instance $x$ | $T$ | Number of top-ranked features in the cumulative weight comparison |
| K | Number of perturbed neighbors generated per instance | $D_R$ | Rank-Weighted Distance between $\phi(x)$ and $\phi(x'_k)$ |
| N | Number of test instances evaluated per model–dataset–condition | $\bar{D}_R$ | Mean Rank-Weighted Distance averaged over K neighbors |
| $\mathcal{D}$ | Dataset of N labeled instances | $D_U$ | Uniform Baseline Distance (equal weights 1/M for all features) |
| $\mathcal{D}_{train}$ | Training partition of the dataset | $\|\phi(x)\|_w$ | Weighted magnitude of the explanation: |

| Symbol | Description | Symbol | Description |
|---|---|---|---|
| $\mathcal{D}_{test}$ | Test partition (stratified 80/20 split) | $\text{CIES}(x)$ | $\sum_j w_j \cdot |\phi_j(x)|$ <br> Credibility Index via Explanation Stability for instance $x$, in $[0, 1]$ |
| $\mathcal{F}_{num}$ | Set of numerical features (subject to perturbation) | $\overline{C}\text{IES}$ | Model-level aggregated CIES: mean ± std over N test instances |
| $\mathcal{F}_{cat}$ | Set of categorical features (held constant under perturbation) | $\|v\|_1$ | L1 norm (Manhattan distance): $\sum_j |v_j|$ |
| $\Phi_f$ | Post-hoc explanation function mapping input $x$ to a feature attribution vector | $\|v\|_2$ | L2 norm (Euclidean distance): $\sqrt{\sum_j v_j^2}$ |
| $\phi(x)$ | Explanation vector — SHAP (or LIME) attributions for instance $x$ | $\|w\|_2$ | L2 norm of the weight vector: $\sqrt{\sum_j w_j^2}$ |
| $\phi_j(x)$ | Attribution of feature $j$ — contribution of feature $j$ to $f(x)$ | $L$ | Local Lipschitz constant of the explanation function at $x$ |
| $\phi(x'_k)$ | Perturbed explanation — attributions recomputed on the $k$-th noisy neighbor | $\hat{L}(x)$ | Estimated local Lipschitz constant |
| $\varepsilon$ | Noise level parameter (e.g. $\varepsilon = 0.03$ means 3% noise) | $\overline{\delta}(x)$ | Mean perturbation magnitude |
| $\eta_j$ | Random noise added to feature $j$, drawn from $\mathcal{N}(0, \sigma_j^2)$ | $Lip_{score}(x)$ | Normalized Lipschitz score: $1 / (1 + \text{Lip}(x))$, in $(0, 1]$ |
| $\sigma_j$ | Standard deviation of noise for feature $j$: $\sigma_j = \varepsilon \cdot |x_j|$ | $\text{PredStab}(x)$ | Prediction stability |
| $\mathcal{N}(0, \sigma_j^2)$ | Gaussian distribution with mean 0 and variance $\sigma_j^2$ | $\rho$ | Spearman rank correlation coefficient |
| $z_k$ | Standard normal random vector: $z_k \sim \mathcal{N}(0, I)$ | $E[D_R]$ | Expected rank-weighted distance over the perturbation distribution |
| $\mathcal{N}_\varepsilon(x)$ | Perturbation neighborhood around $x$ | $p$ | p-value from the Wilcoxon signed-rank test |
| $r_j$ | Rank of feature $j$ when sorted by $|\phi_j(x)|$ descending ($r_j = 1$ = most important) | $\alpha$ | Decay parameter in exponential weighting ($\alpha = 0.5$) |
| $w_j$ | Normalized harmonic weight: $w_j = (1/r_j) / H_M$ | $s$ | SMOTE flag: $s \in \{\text{true}, \text{false}\}$ |